\documentclass[letterpaper, 10 pt, conference]{ieeeconf}  % Comment this line out if you need a4paper

\IEEEoverridecommandlockouts                              % This command is only needed if 
\usepackage[utf8]{inputenc} % allow utf-8 input
\usepackage[T1]{fontenc}    % use 8-bit T1 fonts
\usepackage{inconsolata}

\usepackage{algpseudocode}  
\usepackage[ruled, linesnumbered, boxed]{algorithm2e}
\usepackage{parskip}
\usepackage{changepage}

\usepackage{color}
\usepackage[colorlinks,
            linkcolor=blue,
            anchorcolor=blue,
            citecolor=green,
            urlcolor=Aqua,
            backref=page]{hyperref}
\usepackage{wrapfig}
\usepackage{caption}
\usepackage{subcaption}
\usepackage{cancel}
\usepackage{graphicx}
\usepackage{epsfig} % for postscript graphics files
\usepackage{times} % assumes new font selection scheme installed
\usepackage{booktabs}
\usepackage{booktabs}       % professional-quality tables
\usepackage{amsmath}
\usepackage{amssymb}  % assumes amsmath package installed
\usepackage{dsfont}
\usepackage[table, dvipsnames, svgnames, x11names]{xcolor} 
\usepackage{marvosym}
\definecolor{mygray}{gray}{.9}
\usepackage{pifont}
\newcommand{\cmark}{\ding{51}}%
\newcommand{\xmark}{\ding{55}}%
\usepackage{nicefrac}       % compact symbols for 1/2, etc.
\usepackage{microtype}      % microtypography
\usepackage[table, dvipsnames, svgnames, x11names]{xcolor}        % colors
\usepackage{colortbl}

\usepackage{multirow}
\usepackage{multicol}
\usepackage{cite}
\let\labelindent\relax
\usepackage{enumitem}
\usepackage{graphics} % for pdf, bitmapped graphics files
\usepackage{svg}
\usepackage{bm}
\usepackage{float} 
\usepackage{stfloats}
\usepackage{adjustbox}
\usepackage{tablefootnote}
\usepackage[para]{footmisc}

\newenvironment{sizeddisplay}[1]
{\par\nopagebreak#1\noindent\ignorespaces}
{\nopagebreak\ignorespacesafterend}

\title{\LARGE \bf
H2O+: An Improved Framework for Hybrid Offline-and-Online RL with Dynamics Gaps}

\author{Haoyi Niu$^{1, \spadesuit}$,
Tianying Ji$^{1, \spadesuit}$,
Bingqi Liu$^{2}$,
Haocheng Zhao$^{1}$,
Xiangyu Zhu$^{1}$,\\
Jianying Zheng$^{2}$,
Pengfei Huang$^{1}$,
Guyue Zhou$^{1}$,
Jianming Hu$^{1, \clubsuit}$,
Xianyuan Zhan$^{1,3, \clubsuit}$\thanks{$^1$Tsinghua University. $^2$Beihang University. $^3$Shanghai AI Lab. $^\spadesuit$Equal contribution. 
$^\clubsuit$Corresponding authors. 
This work is supported by National Key Research and Development Program of China under Grant (2022YFB2502904), National Natural Science Foundation of China under Grant 62333015, 62273017, Beijing Natural Science Foundation L231014, and funding from Wuxi Research Institute of Applied Technologies, Tsinghua University under Grant 20242001120.
}
}

\begin{document}

\setlength{\abovedisplayskip}{2pt}
\setlength{\belowdisplayskip}{3pt}
\setlength{\footnotesep}{1pt}

\maketitle
\thispagestyle{empty}
\pagestyle{empty}

% \begin{abstract}
% Heated debates continue about the best autonomous driving framework over the years. Architecture-wise, the modular pipeline is being advantaged in industry owing to its great interpretability and stability, whereas the end-to-end paradigm has demonstrated considerable simplicity and learnability in the rise of deep neural networks. Solutions are therefore expected to combine the superiority of both. ganjue1, it is gradually acknowledged that reinforcement learning overwhelms imitation learning by leaving out the dangerous and costly behavior collecting part in the physical world. Accordingly, we conceive the first modularized end-to-end reinforcement learning framework (ModEL) for autonomous driving to date, a holistic methodology aiming to integrate all these benefits. The human-engineering-economic and sim-to-real-transferable framework offers simple yet effective insights into autonomous driving. Through the examination with both open- and closed-loop metrics, the desirability of this work proves to achieve and even exceed the previous.
% \end{abstract}

% \vspace{-3mm}
\begin{abstract}
Solving real-world complex tasks using reinforcement learning (RL) without high-fidelity simulation environments or large amounts of offline data can be quite challenging. Online RL agents trained in imperfect simulation environments can suffer from severe sim-to-real issues. Offline RL approaches although bypass the need for simulators, often pose demanding requirements on the size and quality of the offline datasets. The recently emerged hybrid offline-and-online RL provides an attractive framework that enables joint use of limited offline data and imperfect simulator for transferable policy learning. In this paper, we develop a new algorithm, called H2O+, which offers great flexibility to bridge various choices of offline and online learning methods, while also accounting for dynamics gaps between the real and simulation environments. Through extensive simulation and real-world robotics experiments,
% \footnote{All the real-world experiment details can be found in \href{https://sites.google.com/view/h2oplusauthors/}{H2O+ webpage}.},
we demonstrate superior performance and flexibility of H2O+ over advanced cross-domain online and offline RL algorithms.

\end{abstract}

\vspace{-1mm}
\section{Introduction}
% \vspace{-1mm}
% Our understanding of the impressive capabilities of Reinforcement Learning (RL) primarily comes from studies on error-tolerant and single-environment tasks such as chess and games. However, a significant challenge arises when applying RL policies across multiple domains with different dynamics. In robotics applications, for instance, a policy learned in a simulator may not translate well to the real-world environment due to differences in dynamics. It has been observed that standard RL algorithms are highly sensitive to changes in environment dynamics, resulting in suboptimal policy performance and limiting the broader success of RL in real-world industrial and everyday-life systems.

% What we know about the superhuman performance of reinforcement learning (RL) is primarily based upon empirical studies that investigate how to handle single-environment tasks like Go~\cite{silver2017mastering} and games~\cite{mnih2015human}
The past successes of reinforcement learning (RL) are primarily restricted to single-domain tasks with the same environment dynamics during the training and testing phases~\cite{silver2017mastering,mnih2015human}.
% largely comes from empirical studies on handling single-environment tasks like Go~\cite{silver2017mastering} and games~\cite{mnih2015human}, commonly with unvaried dynamics in training and deploying phases.
However, it has been observed that most RL algorithms are highly vulnerable to changes in environment dynamics~\cite{luo2022adapt, eysenbach2020off, niu2022when}, resulting in suboptimal policy performance and limiting the broader success of RL in real-world tasks.
In robotics applications~\cite{kober2013reinforcement, lee2020learning, o2022neural, andrychowicz2020learning}, for instance, we typically train control policies in simulators for the sake of training efficiency and safety considerations. However, the dynamics modeling within the simulator can be hard to strictly align with the diverse and complex real-world scenarios, leading to severe performance degradation due to dynamics mismatch~\cite{peng2018sim, sandha2021sim2real, akkaya2019solving}.
% when adapting policies across different domains with dynamics mismatch~\cite{peng2018sim, sandha2021sim2real, akkaya2019solving} .
% transitional dynamics is profoundly entangled in the Markov Decision Process (MDP).
% An intricate issue in reinforcement learning is adapting policies across various domains with dynamics mismatch since transitional dynamics entangles deeply in the Markov Decision Process.
 % As is often the real case in robotics applications~\cite{kober2013reinforcement, lee2020learning, o2022neural, andrychowicz2020learning}, a robot learns policies in simulators, but the underlying dynamics may differ when it is deployed in reality. 
 % It has been observed that standard RL algorithms are highly vulnerable to changes in environment dynamics~\cite{luo2022adapt, eysenbach2020off, niu2022when}, resulting in suboptimal policy performance and limiting the broader success of RL in real-world industrial and everyday-life systems.
% Indeed, existing investigations~\cite{luo2022adapt, eysenbach2020off, niu2022when} have revealed that standard RL algorithms are highly vulnerable and susceptible to the influence of varying environment dynamics, always ending up with the failure to obtain optimal transferable policy performance, which considerably impedes broader success of RL and limits its potential to bring economic benefits to real-world industrial and everyday-life systems.

To address sim-to-real transfer issues~\cite{niu2024comprehensive}, recent RL methods have adopted several design paradigms.
% trends in practical RL paradigm designing have led to a proliferation of studies.
System identification methods~\cite{yu2017preparing,chebotar2019closing, muratore2021data, du2021auto, ramos2019bayessim} aim to calibrate and align the physical properties in simulation with those in the real world. 
Domain randomization techniques~\cite{peng2018sim,rajeswaran2016epopt,mehta2020active, akkaya2019solving} randomize simulation parameters to generalize policies across multiple environments.
% Another line of work incorporates a variety of domain randomization~\cite{peng2018sim,rajeswaran2016epopt,mehta2020active, akkaya2019solving} techniques that randomize the simulation dynamics parameters to generalize policies across multiple environments
% , with the expectation that they will also apply to the real environment. 
However, the selection of parameters and the range of their randomization could require a great amount of human effort and domain expertise~\cite{vuong2019pick, andrychowicz2020learning}, as well as sufficient configurability of the simulator~~\cite{chen2022real}. 
% Furthermore, not all the simulations are differentiable and controllable for accessing and even altering their dynamics modeling~\cite{chen2022real} . 
Thus, another avenue of works~\cite{eysenbach2020off,liu2021dara} regard simulators as black boxes and turn to perform policy learning adaptation via modifying the reward to account for the sim-to-real dynamics gap. 
More recently, the rapid developments in offline RL~\cite{levine2020offline,fujimoto2019off,kumar2019stabilizing,fujimoto2021minimalist,kostrikov2022offline,por,xu2023offline,garg2023extreme} have brought renewed interest in learning policies directly from pre-collected real-world datasets to bypass the need for simulation environments. These methods adopt conservative principles to overcome the notorious distributional shift issue~\cite{kumar2019stabilizing} in offline learning, thus often requiring large, high state-action space coverage, and high-quality datasets to achieve good performance~\cite{li2023data} which can be hard to satisfy in scenarios with high data collection costs.
% in case samples form incorrect simulation dynamics would contaminate the learning process. 
% Nevertheless, offline RL tailors multiple conservative learning formalisms to overcome the notorious distribution shift issues~\cite{kumar2019stabilizing} that also limit their potentials in turn. Plus, offline datasets obtained from the real world often suffer from issues such as poor quality, limited quantity, or narrow coverage due to the high costs associated with data collection efforts.
% offline datasets acquired from the real world are often of low quality, insufficient quantity or narrow coverage due to costly labor efforts.
% ~\cite{yu2017preparing,chebotar2019closing,zhouenvironment,du2021auto}

\vspace{-1.5mm}
All previous approaches 
% All of the aforementioned approaches 
bear some limitations, suggesting that solely relying on online simulation samples with imperfect dynamics or limited, low-coverage real-world offline data may not be sufficient to achieve desirable policy transferability.
% Therefore, policies learned from online simulation samples with imperfect dynamics or offline real-world data, respectively, can hardly perform with the expected transferability. 
To this end, dynamics-aware hybrid offline-and-online RL (H2O)~\cite{niu2022when} is the first study to combine offline and online policy learning using both limited offline real-world data and off-dynamics online simulated samples for cross-domain policy learning.
It introduces a dynamics-aware value regularization scheme that boosts Q-values on offline real data and punishes Q-values on simulation samples based on explicit dynamics gap quantification. 
% It proves to be overwhelming online zero-transfer RL, offline RL, and other cross-domain methods that also leverage offline and online data with dynamics gaps. 
% especially in real robot situations.
Although promising, H2O bears several drawbacks. First, it is built on the conservative Q-learning (CQL)~\cite{kumar2020conservative} framework, which is over-conservative and lacks flexibility to be extended to stronger and less conservative offline RL paradigms. Its over-conservative design hinders sufficient exploration and state-action coverage improvement in the simulation environment. Lastly, explicit dynamics gap quantification in H2O also poses computation challenges.

% However, central to the entire discipline of H2O is an over-conservative value regularization backbone of CQL~\cite{kumar2020conservative}, which lacks flexibility for extension to stronger and less conservative offline RL paradigms. Moreover,
% % a waste of the original intention to incorporate online interactions into offline learning is that 
% H2O actually makes no attempt to encourage thorough exploration in simulators. Additionally, dynamics gap estimation in H2O bears numerical approximations that could sometimes cause misleading value distortion at state-action pairs without sufficient data coverage, as well as costly computational burdens for implementations.
\vspace{-1mm}
In this paper, we follow the offline-and-online RL recipe in H2O, but develop a more flexible and powerful algorithm through a different lens, to enable sufficient utilization of both the offline dataset and imperfect simulator for transferable policy learning. We refer our algorithm as \textbf{H2O+}, which has two favorable design ingredients: 1) a \textit{flexible and less conservative learning framework} that is compatible with various strong in-sample learning offline RL backbones and exploration designs; and 2) the \textit{dynamics-aware mixed value update} that bridges offline and online value function learning, while also accounting for dynamics gaps between real and simulated samples.
% In consideration of those drawbacks, we are motivated to rethink the dynamics-aware offline-and-online recipe (\textbf{H2O+}) with more desirable design ingredients:
% (1) \textbf{a flexible and less conservative learning framework} that can be compatible with various offline backbone and exploration choices.
% (2) \textbf{dynamics-aware mixed value update} that linearly combines reliable state value function $V(\mathbf{s})$ learned from behavior-regularized offline RL and target network $Q(\mathbf{s},\mathbf{a})$ encouraged by maximum entropy RL as a new form of target values, marrying the benefits from both as well as addressing dynamics gap. 
% % with tuning the learning signals for Q-values on simulation samples. 
% On one hand, an optimistic $Q(\mathbf{s},\mathbf{a})$ would guide policies to perform full exploration in simulators; and on the other hand, $V(\mathbf{s})$ learned solely on offline data would stabilize the value learning and capture the high-quality data among them, also serving as an anchor to mildly regulate $Q(\mathbf{s},\mathbf{a})$ estimation on potentially biased simulation samples. 
% Plus fixing the bellman error with dynamics ratio, H2O+ reaches a subtly balance that we not only unleash the potential for simulation exploration to the utmost but also avoid biased simulation data to disturb full utilization of offline real data. 
Through extensive simulation and real wheel-legged robot experiments, we demonstrate the superiority and flexibility of H2O+ over competing online, offline, and cross-domain RL baseline methods.

\vspace{-2mm}
\section{Related Work}\label{rw}
\vspace{-2mm}

\subsection{Reinforcement Learning with Imperfect Simulators}
\vspace{-2mm}
High-fidelity simulators are crucial for online RL methods to learn deployable policies. However, as accurate simulators are hard to build, addressing the sim-to-real gaps become a pressing challenge. Various cross-domain online RL approaches have been proposed to tackle this challenge, such as using system identification methods~\cite{ljung1998system, chebotar2019closing, muratore2021data, du2021auto, ramos2019bayessim} to align simulated dynamics with real dynamics, or adding domain randomizations~\cite{peng2018sim, rajeswaran2016epopt, andrychowicz2020learning, mehta2020active, akkaya2019solving} that trains RL policies in a randomized simulated dynamics setting. The former typically requires a considerable amount of offline or costly real-world interaction data~\cite{yu2017preparing}, while the latter necessitates manually-specified randomized parameters~\cite{vuong2019pick}. Recently, another line of research leverages additional real-world data to mitigate the dynamics shift in simulation environments~\cite{eysenbach2020off,liu2021dara,niu2022when}. Specifically, DARC~\cite{eysenbach2020off} and DARA~\cite{liu2021dara} add dynamics-gap-related penalization terms on rewards in online and offline RL settings, respectively. H2O~\cite{niu2022when} proposes a new setting that enables simultaneous offline-and-online policy learning on both real offline data and simulated samples, which shows promising results and advantages over prior methods. 

% Addressing sim2real dynamics gaps is a long-standing challenge. 
% Various online RL methods have been proposed, from the use of system identification methods~\cite{ljung1998system, chebotar2019closing, muratore2021data, du2021auto, ramos2019bayessim} that align simulated dynamics with real-world dynamics to Domain Randomization~(DR)~\cite{peng2018sim, rajeswaran2016epopt, andrychowicz2020learning, mehta2020active, akkaya2019solving} that trains RL policies in a randomized simulated dynamics setting. 
% While the former requires considerable offline or real-world interaction data~\cite{yu2017preparing}, which can be costly and sometimes impractical, the latter necessitates manually-specified randomized parameters~\cite{vuong2019pick}.
% Another line of work, Dynamics Adaptation, \emph{e.g.},~\cite{eysenbach2020off}, demands both real and simulated trajectories for training two discriminators, which is not only expensive but also requires unlimited real-world interactions.
% \zhan{H2O}

\vspace{-5pt}
\subsection{Policy Learning by Combining Offline and Online RL}
The recently emerged offline RL methods~\cite{fujimoto2019off,kumar2019stabilizing,fujimoto2021minimalist,kostrikov2022offline,por,xu2023offline,garg2023extreme} has provided an attractive solution to learn policies directly from offline data without online interactions. However, the performances of existing offline RL methods are heavily limited by the quality and state-action space coverage of offline datasets~\cite{kumar2019stabilizing,li2023data}. To mitigate this issue, offline-to-online RL methods~\cite{nair2020awac,lee2022offline,zhang2023policy} are developed to separate RL policy learning into a two-stage training process: first pretrain a policy using offline RL and then finetune with online RL. It can improve sample efficiency with favorable initialization for the online learning stage. More recently, some RL studies~\cite{song2023hybrid,ball2023efficient,wagenmaker2022leveraging,ji2023seizing} directly merge offline RL ingredients into online RL algorithms as a single-stage learning process, which have been shown to greatly improve sample efficiency and policy performance. However, all these methods are only applicable to a single domain, with no dynamics gaps between the online and offline data. H2O~\cite{niu2022when} also adopts simultaneous offline and online learning, but is specifically designed to tackle off-dynamics online samples from an imperfect simulator. Our proposed H2O+ follows the same hybrid offline-and-online RL setting, but uses a different methodological framework to address several key drawbacks of H2O.

\section{Preliminaries}
% \subsection{Reinforcement Learning}
\subsection{Reinforcement Learning}
\vspace{-4pt}
% \zhan{MDP}
We consider the RL problem formulated as a Markov Decision Process~(MDP)~\cite{sutton1998introduction}, defined by a tuple $\mathcal{M}:=\left(\mathcal{S}, \mathcal{A}, r, P_{\mathcal{M}}, \gamma\right)$. $\mathcal{S}$ and $\mathcal{A}$ denote the state and action space, $r$ represents the reward function, $P_{\mathcal{M}}$ stands for the transition dynamics under $\mathcal{M}$. 
The goal of RL is
to find the optimal policy $\pi^*$ that maximizes cumulative discounted reward starting from an initial state distribution $\rho$, \emph{i.e.}, 
% \begin{displaymath}
%     \pi^* = \arg\max_{\pi} \mathbb{E}_{\mathbf{s}_0\in\rho,\mathbf{a}_t\sim\pi(\cdot|\mathbf{s}_t), \mathbf{s}_{t+1}\sim P_{\mathcal{M}}(\cdot|\mathbf{s}, \mathbf{a})}\left[\sum_{t=0}^{\infty} \gamma^{t} r\left(\mathbf{s}_{t}, \mathbf{a}_{t}\right)\right]
% \end{displaymath}
$\pi^* =\arg\max_{\pi} \mathbb{E}_{\mathbf{s}_0\in\rho,\mathbf{a}_t\sim\pi(\cdot|\mathbf{s}_t), \mathbf{s}_{t+1}\sim P_{\mathcal{M}}(\cdot|\mathbf{s}, \mathbf{a})}\left[\sum_{t=0}^{\infty} \gamma^{t} r\left(\mathbf{s}_{t}, \mathbf{a}_{t}\right)\right]$. RL methods based on approximated dynamic programming typically learn an action-value function $Q(s,a)$, and optionally, a state value function $V(s)$ to practically estimate the cumulative discounted reward for policy optimization.

In many cases, RL training in the real environment is infeasible, so most online RL methods train the agents in simulation environments. However, building a high-fidelity simulator can be costly or even impossible in many real-world tasks. Learning with an imperfect simulator will lead to a MDP $\widehat{\mathcal{M}}:=\left(\mathcal{S}, \mathcal{A}, r, P_{\widehat{\mathcal{M}}}, \gamma\right)$ with biased dynamics $P_{\widehat{\mathcal{M}}}$, which can cause serious sim-to-real transfer issues. When a large offline real-world dataset $\mathcal{D}$ generated by some behavior policy $\mu$ is given, one can also resort to offline RL~\cite{fujimoto2019off, levine2020offline} to bypass the sim-to-real issue and directly learn a policy from the offline data. However, the performances of existing offline RL methods are heavily dependent on the size and quality of datasets, which restricts their practical application~\cite{li2023data}.

% \vspace{-8pt}
% \zhan{introduce the symbol $\mu$ as behavior policy}
\subsection{Hybrid Offline-and-Online RL with Imperfect Simulator}
% \vspace{-4pt}
% \zhan{H2O, one equations}
As both online and offline RL bear some practical challenges in solving real-world problems, there is a growing interest in merging online and offline RL for sample-efficient and high-performance policy learning~\cite{song2023hybrid,ball2023efficient,wagenmaker2022leveraging,niu2022when}. Many of these studies~\cite{song2023hybrid,ball2023efficient,wagenmaker2022leveraging} assume identical online and offline system dynamics, thus are not applicable if an imperfect simulator is used as the online environment. 
% There has been recent interest in hybrid offline-and-online RL settings, where an agent is provided with an offline dataset and has access to an online simulator. Most of them~\cite{song2023hybrid,ball2023efficient}, however, rely on a perfect simulator with no dynamic gaps, which is often unattainable in real-world situations. 
Among them, H2O~\cite{niu2022when} is the first study that enables simultaneous offline and online policy learning with an imperfect simulator. H2O is built upon the conservative Q-learning (CQL)~\cite{kumar2020conservative} framework, with its learning objective designed as follows:
% applies a dynamics-gap weighted value regularization on simulated samples 
% for value regularization, and the second representing the Bellman error of mixed data from the offline dataset $\mathcal{D}$ and the simulation rollout samples in the online replay buffer $B$, state as follows
\begin{sizeddisplay}{\scriptsize}
\begin{align}
    &\min_{Q}\; \underbrace{\alpha_c\cdot\Big(\log \sum_{\mathbf{s},\mathbf{a}} {\omega(\mathbf{s},\mathbf{a})} \exp \left(Q(\mathbf{s},\mathbf{a})\right)-\mathbb{E}_{\mathbf{s},\mathbf{a}\sim \mathcal{D}}\left[Q(\mathbf{s},\mathbf{a})\right]\Big)}_{\textit{\footnotesize (i) Conservative value regularization}}+ \label{eq:h2o}\\
    &\underbrace{\mathbb{E}_{\mathcal{D}}\left[\left(Q-\hat{\mathcal{B}}^{\pi} \hat{Q}\right)(\mathbf{s},\mathbf{a})\right]^{2}+\mathbb{E}_{B}\frac{P_{\mathcal{M}}(\mathbf{s}' | \mathbf{s}, \mathbf{a})}{P_{\widehat{\mathcal{M}}}(\mathbf{s}' | \mathbf{s}, \mathbf{a})}\left[\left(Q-\hat{\mathcal{B}}^{\pi} \hat{Q}\right)(\mathbf{s},\mathbf{a})\right]^{2}}_{\textit{\footnotesize (ii) Bellman error on offline and online data}}\notag
\end{align}
\end{sizeddisplay}
% \begin{sizeddisplay}{\scriptsize}
% \begin{align}
%     &\min_{Q}\; \underbrace{\alpha_c\cdot\Big(\log \sum_{\mathbf{s},\mathbf{a}} {\omega(\mathbf{s},\mathbf{a})} \exp \left(Q(\mathbf{s},\mathbf{a})\right)-\mathbb{E}_{\mathbf{s},\mathbf{a}\sim \mathcal{D}}\left[Q(\mathbf{s},\mathbf{a})\right]\Big)}_{\textit{\footnotesize (i) Conservative value regularization}}+ \label{eq:h2o}\\
%     &\underbrace{\mathbb{E}_{(\mathbf{s}, \mathbf{a},\mathbf{s}')\sim \mathcal{D}\cup  B}\frac{P_{\mathcal{M}}(\mathbf{s}' | \mathbf{s}, \mathbf{a})}{P_{\widehat{\mathcal{M}}}(\mathbf{s}' | \mathbf{s}, \mathbf{a})}\Bigg|_{(\mathbf{s}, \mathbf{a},\mathbf{s}')\sim B}\left[\left(Q-\hat{\mathcal{B}}^{\pi} \hat{Q}\right)(\mathbf{s},\mathbf{a})\right]^{2}}_{\textit{\footnotesize (ii) Bellman error on offline and online data}}\notag
% \end{align}
% \end{sizeddisplay}
H2O's learning objective is comprised of two parts: the first part pushes down dynamics-gap weighted Q-values and pulls up Q-values on trustworthy real offline data; the second part enables simultaneous offline and online learning on both offline dataset $\mathcal{D}$ and simulated replay buffer $B$ while also correcting the problematic next state $\mathbf{s}'$ from the simulator dynamics $P_{\widehat{\mathcal{M}}}$ using the dynamics ratio as an importance weight. In H2O, the dynamics gap measure $\omega(\mathbf{s},\mathbf{a})$ is explicitly calculated as the normalized KL-divergence $D_{KL}(P_{\widehat{\mathcal{M}}}(\mathbf{s}'|\mathbf{s},\mathbf{a})\|P_{\mathcal{M}}(\mathbf{s}'|\mathbf{s},\mathbf{a}))$ over all $(\mathbf{s},\mathbf{a})$ pairs in the state-action space, which can only be approximated.

\section{Method}\label{method}
Although H2O provides a successful attempt to tackle sim-to-real dynamics gaps by combining offline and online RL, it suffers from four notable drawbacks. First, the CQL backbone of H2O is over-conservative~\cite{nakamoto2023cal,li2023data} and may cause conflict when incorporating online learning. For example, as shown in~\cite{nakamoto2023cal}, when performing online fine-tuning on a conservative value function initialization, policy learning has to first "unlearn" the underestimated values before making further progress. Second, the CQL framework lacks flexibility, which is not possible to be extended nor compatible with many recent strong offline RL frameworks~\cite{kostrikov2022offline,xu2023offline,hansen2023idql,garg2023extreme}. H2O also has no exploration design, which hinders effective state-action coverage improvement through simulation interactions. Lastly, H2O has to approximate an explicit dynamics gap measure, which is costly and error-prone.
These drawbacks motivate us to rethink what are the desirable properties in hybrid offline-and-online RL. In this paper, we propose H2O+, which offers a highly flexible and less conservative algorithm, enabling full utilization of the online samples from the imperfect simulator. The key of H2O+ is the dynamics-aware mixed value update, which bridges various choices of offline and online learning methods, while also accounting for dynamics gaps between the real and simulation environment.

% In order to address the dynamics mismatch, H2O utilizes a value regularization strategy to lower the Q-values on those simulation samples in accordance with dynamics gap quantification, however, sometimes leading to severe over-conservatism. In fact, value regularization manages dynamics shift as well as distribution shift, which seems superfluous in the hybrid situation with ongoing updates from online simulation. We recommend the most essential ingredient is to perform in-dynamics exploration sufficiently in simulation. 

\subsection{Separate Considerations for Offline and Online Learning}\label{subsec:off-on}
Before introducing our method, we first review two popular modeling frameworks in both offline and online RL: behavior-regularized RL (Eq.(\ref{eq:offline_J})) and maximum entropy RL~(Eq.(\ref{eq:online_J})):
% \begin{equation}
% \small
%     \text{Offline:} \  \max_\pi
% \mathbb{E}\left[\sum_{t=0}^{\infty} \gamma^t\left(r\left(\mathbf{s}_t, \mathbf{a}_t\right)-\alpha\cdot f\left( \frac{\pi\left(\mathbf{a}_t \mid \mathbf{s}_t\right)}{\mu\left(\mathbf{a}_t \mid \mathbf{s}_t\right)}\right)\right)\right] \label{eq:offline_J}
% \end{equation}
% \begin{equation}
% \small
%     \text{Online:} \    \max_\pi \mathbb{E}\left[\sum_{t=0}^{\infty} \gamma^t\left(r\left(\mathbf{s}_t, \mathbf{a}_t\right)+\beta\cdot \mathcal{H}\left(\pi\left(\mathbf{a}_t \mid \mathbf{s}_t\right)\right)\right)\right]\label{eq:online_J}
% \end{equation}
% \begin{equation}
% \small
%     \text{Offline:} \;  \max_\pi
% \mathbb{E}\left[\sum_{t=0}^{\infty} \gamma^t\left(r\left(\mathbf{s}_t, \mathbf{a}_t\right)-\alpha\cdot f\left( \frac{\pi\left(\mathbf{a}_t \mid \mathbf{s}_t\right)}{\mu\left(\mathbf{a}_t \mid \mathbf{s}_t\right)}\right)\right)\right] \label{eq:offline_J}
% \end{equation}
% \vspace{6pt}
% \begin{equation}
% \small
%     \text{Online:} \;    \max_\pi \mathbb{E}\left[\sum_{t=0}^{\infty} \gamma^t\left(r\left(\mathbf{s}_t, \mathbf{a}_t\right)+\beta\cdot \mathcal{H}\left(\pi\left(\mathbf{a}_t \mid \mathbf{s}_t\right)\right)\right)\right]\quad\;\label{eq:online_J}
% \end{equation}
\begin{sizeddisplay}{\small}
\begin{align}
 \text{Offline:}& \;  \max_\pi
\mathbb{E}\left[\sum_{t=0}^{\infty} \gamma^t\left(r\left(\mathbf{s}_t, \mathbf{a}_t\right)-\alpha\cdot f\left( \frac{\pi\left(\mathbf{a}_t \mid \mathbf{s}_t\right)}{\mu\left(\mathbf{a}_t \mid \mathbf{s}_t\right)}\right)\right)\right] \label{eq:offline_J}\\
  \text{Online:}& \;    \max_\pi \mathbb{E}\left[\sum_{t=0}^{\infty} \gamma^t\left(r\left(\mathbf{s}_t, \mathbf{a}_t\right)+\beta\cdot \mathcal{H}\left(\pi\left(\mathbf{a}_t \mid \mathbf{s}_t\right)\right)\right)\right]\label{eq:online_J}
\end{align}
\end{sizeddisplay}
The behavior-regularized RL is formally studied in~\cite{xu2023offline}, which has been shown closely related to a class of recent state-of-the-art (SOTA) in-sample learning offline RL methods~\cite{xu2023offline,hansen2023idql}. 
% {\color{blue} These methods learn improved value functions completely using samples in the offline dataset and then extract the policy using the learned value function. The entire learning process only involves samples in the offline datasets, rather than policy-induced actions.
% , hence is referred to as "in-sample learning" methods.
% }
Depending on different choices of the $f$ function, it can be shown that all these algorithms share the following general learning objectives for 
% the state- and action-value functions 
$V(\mathbf{s})$ and $Q(\mathbf{s},\mathbf{a})$:
\begin{sizeddisplay}{\small}
\begin{align}
    &\min_V \mathbb{E}_{(\mathbf{s},\mathbf{a})\sim \mathcal{D}} \;\mathcal{L}_V^f(Q(\mathbf{s},\mathbf{a})-V(\mathbf{s})) \label{eq:insample_V}\\
    &\min_Q \;\mathbb{E}_{(\mathbf{s},\mathbf{a},\mathbf{s}')\sim \mathcal{D}}\left[ r(\mathbf{s},\mathbf{a}) + \gamma V(\mathbf{s}')- Q(\mathbf{s},\mathbf{a})\right]^2
    % =\mathbb{E}_{(\mathbf{s},\mathbf{a},s')\sim \mathcal{D}}\left[ \mathcal{B}_{\text{off}}Q(\mathbf{s},\mathbf{a})- Q(\mathbf{s},\mathbf{a})\right]^2
    \label{eq:insample_Q}
\end{align}
\end{sizeddisplay}
% where $\mathcal{B}_{\text{off}}Q(\mathbf{s},\mathbf{a})=r(\mathbf{s},\mathbf{a}) + \gamma V(s')$. 
In particular, if $f=\log(x)$, it correspond to EQL~\cite{xu2023offline} and XQL~\cite{garg2023extreme} with $\mathcal{L}_V^f(y)=\exp(y/\alpha)-y/\alpha$. If $f=x-1$, it corresponds to SQL~\cite{xu2023offline} (equivalent to an in-sample learning version of CQL) with $\mathcal{L}_V^f(y)=\mathds{1}(1+y/2\alpha>0)(1+y/2\alpha)^2-y/\alpha$. The well-known offline RL algorithm IQL~\cite{kostrikov2022offline} also belongs to this family of algorithms but does not have a closed-form $f$, with $\mathcal{L}_V^f(y)=|\tau-\mathds{1}(y<0)| y^2$, where $\tau\in (0,1)$ is the expectile hyperparameter.
These offline RL methods learn $V(\mathbf{s})$ and $Q(\mathbf{s},\mathbf{a})$ completely using dataset samples, thus enjoying stable value function learning as compared to CQL-style algorithms that distort the value estimates. However, their in-sample learning nature also creates obstacles to incorporating online learning with imperfect simulators.

On the other hand, the maximum entropy RL~\cite{haarnoja2018soft}~(Eq.(\ref{eq:online_J})) also achieves great success in online RL studies, which maximizes the expected reward while also maximizing the entropy of the policy $\mathcal{H(\pi})$ to promote exploration. If we consider an off-policy setting and denote $B$ as the training replay buffer, its corresponding action-value function learning objective is given as:
\begin{equation}
\small
\begin{aligned}
    \min_Q \;\mathbb{E}_{(\mathbf{s},\mathbf{a})\sim B}\big[ &r(\mathbf{s},\mathbf{a}) + \gamma \mathbb{E}_{\mathbf{s}'\sim P, \mathbf{a}'\sim \pi} \big[\hat{Q}(\mathbf{s}',\mathbf{a}') \\
    &-\beta\cdot\log(\pi(\mathbf{a}'|\mathbf{s}'))\big]- Q(\mathbf{s},\mathbf{a})\big]^2 \label{eq:evaluation_Q}
    % \\    &=\mathbb{E}_{(\mathbf{s},\mathbf{a})\sim B}\left[ B^{\pi}_{\text{on}}Q(\mathbf{s},\mathbf{a})- Q(\mathbf{s},\mathbf{a})\right]^2
    \end{aligned}
\end{equation}
% The largest advantage of leveraging simulators is the capability for broad and unlimited exploration. The general optimization objective for explorative learning can be given by:
% \begin{equation}
% J(\pi)=\mathbb{E}_{\mathbf{a} \sim \pi}\left[\sum_{t=0}^{\infty} \gamma^t\left(r\left(\mathbf{s}_t, \mathbf{a}_t\right)-\beta g\left(\pi\left(\mathbf{a}_t \mid \mathbf{s}_t\right)\right)\right)\right]
% \end{equation}
% where we can define $g$ with diverse choices of exploration terms: entropy function~\cite{haarnoja2018soft}, guassian noise~\cite{fujimoto2018addressing} and pink noise~\cite{eberhard2023pink}.
\subsection{Dynamics-Aware Mixed Value Update}\label{sec:mix}
Both the behavior-regularized RL and the maximum entropy RL frameworks bear some attractive features for the hybrid offline-and-online RL setting. For example, the behavior-regularized RL provides high-quality offline learned value functions without posing too much conservatism. While the maximum entropy RL offers natural exploration capabilities to help improve the state-action space coverage of the offline dataset. Now the question is: how can we leverage the strengths of the two frameworks to build a strong hybrid RL algorithm, while also being capable of tackling the sim-to-real dynamics gaps between real and simulation environments?

In this paper, we provide a simple and elegant solution by proposing a dynamics-aware mixed value update that seamlessly mixes offline and online learning without introducing excessive conservatism. Our insight is by noting that we can use the more reliable state value function $V(\mathbf{s})$ learned solely with the real offline data in Eq.(\ref{eq:insample_V}) as an anchor to mildly regulate $Q(\mathbf{s}, \mathbf{a})$ estimation on potentially biased simulation samples. We can achieve this by utilizing the following mixed Bellman operator~\cite{ji2023seizing}, balancing exploitation to real-world offline data and exploration with online simulation data:
\begin{sizeddisplay}{\small}
\begin{align}
&\mathcal{B}^{\text{mix}}_{\lambda} \hat{Q}(\mathbf{s},\mathbf{a})=\lambda \big[r(\mathbf{s},\mathbf{a})+\gamma V(\mathbf{s}^{\prime})\big] + \notag\\
&(1-\lambda) \left[ r(\mathbf{s},\mathbf{a}) +
\gamma \mathbb{E}_{\mathbf{a}'\sim \pi} \left[\hat{Q}(\mathbf{s}',\mathbf{a}')-\beta\cdot\log(\pi(\mathbf{a}'|\mathbf{s}'))\right]\right] \label{eq:mix}\\
&=r(\mathbf{s}, \mathbf{a}) + \lambda \gamma V(\mathbf{s}^{\prime})+(1-\lambda) \gamma \mathbb{E}_{\pi} \left[\hat{Q}(\mathbf{s}',\mathbf{a}')-\beta\cdot\log\pi(\mathbf{a}'|\mathbf{s}')\right]\notag
\end{align}
\end{sizeddisplay}
where the state value function is learned only with real-world offline data $\mathcal{D}$ as in Eq.(\ref{eq:insample_V}); $\lambda\in [0,1]$ is a trade-off hyperparameter to control the level of influence between offline and online learning. With the mixed Bellman operator, we can learn Q-function from both real dataset $\mathcal{D}$ and online simulation replay buffer $B$. Moreover, to correct potentially problematic next states $\mathbf{s}'$ from the simulator dynamics $P_{\widehat{\mathcal{M}}}$, we adopt the same dynamics ratio reweighting as in H2O~\cite{niu2022when}:
\begin{equation}
\small
    \begin{aligned}
  \min_{Q}\; &\mathbb{E}_{\left(\mathbf{s}, \mathbf{a}, \mathbf{s}^{\prime}\right) \sim \mathcal{D}}\left[\left(Q-\mathcal{B}^{\text{mix}}_{\lambda} \hat{Q}\right)^2(\mathbf{s}, \mathbf{a})\right] + \\
  &\mathbb{E}_{\left(\mathbf{s}, \mathbf{a}\right) \sim B} \mathbb{E}_{\mathbf{s}^{\prime}\sim P_\mathcal{M}} \left[\left(Q-\mathcal{B}^{\text{mix}}_{\lambda} \hat{Q}\right)^2(\mathbf{s}, \mathbf{a})\right] \\
  =&\mathbb{E}_{\left(\mathbf{s}, \mathbf{a}, \mathbf{s}^{\prime}\right) \sim \mathcal{D}}\left[\left(Q-\mathcal{B}^{\text{mix}}_{\lambda} \hat{Q}\right)^2(\mathbf{s}, \mathbf{a})\right] + \\
  &\mathbb{E}_{\left(\mathbf{s}, \mathbf{a}, \mathbf{s}^{\prime}\right) \sim B}\left[\frac{P_{\mathcal{M}}\left(\mathbf{s}^{\prime} \mid \mathbf{s}, \mathbf{a}\right)}{P_{\widehat{\mathcal{M}}}\left(\mathbf{s}^{\prime} \mid \mathbf{s}, \mathbf{a}\right)}\left(Q-\mathcal{B}^{\text{mix}} _{\lambda}\hat{Q}\right)^2(\mathbf{s}, \mathbf{a})\right]\label{eq:final_Q}
\end{aligned}
\end{equation}
The dynamics ratio $P_{\widehat{\mathcal{M}}}/P_{\mathcal{M}}$ can be conveniently estimated by learning a pair of domain discriminators $p(\text{real}|\mathbf{s}, \mathbf{a}, \mathbf{s}^{\prime})$ and $p(\text{real}|\mathbf{s}, \mathbf{a})$ using the following formulation, which is also adopted in a number of previous studies~\cite{eysenbach2020off,liu2021dara,niu2022when}: 
% where we estimate the dynamics ratio in Eq.~(\ref{eq:final_Q}) with a pair of domain discriminators $D_{sas}$ and $D_{sa}$ that approximate $p\left(\text {real} \mid \mathbf{s}, \mathbf{a}, \mathbf{s}^{\prime}\right)$ and $p\left(\text {real} \mid \mathbf{s}, \mathbf{a}\right)$ respectively in the below formulation~\cite{eysenbach2020off,liu2021dara,niu2022when}
\begin{sizeddisplay}{\small}
\begin{align}
\frac{P_{\mathcal{M}}\left(\mathbf{s}^{\prime} |\mathbf{s}, \mathbf{a}\right)}{P_{\widehat{\mathcal{M}}}\left(\mathbf{s}^{\prime}|\mathbf{s}, \mathbf{a}\right)}
&=\frac{p\left(\mathbf{s}^{\prime}|\mathbf{s}, \mathbf{a}, \text{real}\right)}{p\left(\mathbf{s}^{\prime}|\mathbf{s}, \mathbf{a}, \text{sim}\right)}
=\frac{p(\text{sim}|\mathbf{s}, \mathbf{a})}{p(\text {real}|\mathbf{s}, \mathbf{a})} / \frac{p\left(\text{sim}|\mathbf{s}, \mathbf{a}, \mathbf{s}^{\prime}\right)}{p\left(\text{real}|\mathbf{s}, \mathbf{a}, \mathbf{s}^{\prime}\right)} \notag\\
&= \frac{1-p(\text{real}|\mathbf{s}, \mathbf{a})}{p(\text{real}|\mathbf{s}, \mathbf{a})} / \frac{1-p\left(\text{real}|\mathbf{s}, \mathbf{a}, \mathbf{s}^{\prime}\right)}{p\left(\text{real}|\mathbf{s}, \mathbf{a}, \mathbf{s}^{\prime}\right)}\label{eq:dr}
\end{align}
\end{sizeddisplay}
% The discriminators are optimized with cross-entropy loss:
% \begin{equation}
% \begin{aligned}
% &D_{sas}^* \leftarrow \arg\max_{D_{sas}} -\mathbb{E}_{\left(\mathbf{s}, \mathbf{a}, \mathbf{s}^{\prime}\right) \sim \mathcal{D}}\left[\log D_{sas}\left(\mathbf{s}, \mathbf{a}, \mathbf{s}^{\prime}\right)\right] -\mathbb{E}_{\left(\mathbf{s}, \mathbf{a}, \mathbf{s}^{\prime}\right) \sim B}\left[\log \left(1-D_{sas}\left(\mathbf{s}, \mathbf{a}, \mathbf{s}^{\prime}\right)\right)\right]\\
% &D_{sa}^* \leftarrow \arg\max_{D_{sa}} -\mathbb{E}_{\left(\mathbf{s}, \mathbf{a}\right) \sim \mathcal{D}}\left[\log D_{sa}\left(\mathbf{s}, \mathbf{a}\right)\right] -\mathbb{E}_{\left(\mathbf{s}, \mathbf{a}\right) \sim B}\left[\log \left(1-D_{sa}\left(\mathbf{s}, \mathbf{a}\right)\right)\right]\label{dis_loss}
% \end{aligned}
% \end{equation}
Finally, with the learned action value function $Q(\mathbf{s}, \mathbf{a})$, we can optimize the policy $\pi$ by maximizing the following objective on both real and simulated samples:
% Finally, we extract the policy by maximizing the following objective:
\begin{equation}
\small
\pi^* = \arg\max_{\pi} \mathbb{E}_{(\mathbf{s}, \mathbf{a})\in\mathcal{D}\cup B} [Q(\mathbf{s}, \mathbf{a})-\beta\cdot\log(\pi(\mathbf{s}, \mathbf{a}))]
\end{equation}

\subsection{Discussion and Comparison with H2O} 
Using the above dynamics-aware mixed value update, H2O+ effectively addresses all the aforementioned drawbacks of H2O. First, H2O+ uses in-sample learning state-value function $V(s)$ and the dynamics ratio to mildly regulate the value function learning on potentially problematic online simulated samples. There is no distortion nor extra conservative penalty on the Q-values, thus removing excessive conservatism during policy learning. Second, H2O+ is compatible with a series of recent strong in-sample learning offline RL methods~\cite{kostrikov2022offline,xu2023offline,hansen2023idql,garg2023extreme}, and the policy entropy in Eq.(\ref{eq:mix}) is also possible to be replaced with other terms to promote exploration, thus offering great flexibility. Moreover, H2O+ removes the need to estimate explicit dynamics gap measures, thus providing a simpler and more efficient algorithmic implementation. In the next section, we will show in empirical experiments, that although it removes much conservatism to regulate off-dynamics samples, H2O+ consistently outperforms H2O and other cross-domain RL baselines.

 \section{Experiments}\label{exp}

In this section, we present empirical validations of our approach. 
We begin with our algorithmic implementation and experimental setups, followed by benchmark experiments with original and dynamics-modified MuJoCo simulation environments. 
% with intentionally introducing multiple types of dynamics gap in original MuJoCo simulation. 
Our baselines consist of the online RL method SAC~\cite{haarnoja2018soft} for zero-shot transfer, offline RL algorithms CQL~\cite{kumar2020conservative} and IQL~\cite{kostrikov2022offline}, cross-domain online RL method DARC~\cite{eysenbach2020off}, and H2O~\cite{niu2022when} that in a similar setting with H2O+. We run all experiments with 5 random seeds.
Finally, we deploy H2O+ and baselines on a wheel-legged robot to complete real-world tasks.
Furthermore, we provide ablations on choices of dynamics-aware mixed value update designings, different levels of intensity of dynamics gap and different offline RL backbones.
\subsection{Experimental Setups}\label{setup}
\subsubsection{Algorithmic implementation of H2O+} In all our comparative experiments, we instantiate $\mathcal{L}_V^f(y)=|\tau-\mathds{1}(y<0)| y^2$ as in IQL~\cite{kostrikov2022offline}, due to its simplicity. The scaling parameter $\beta$ of the entropy term is automatically tuned following the treatment in SAC~\cite{haarnoja2018soft}.
We follow the treatment in SAC~\cite{haarnoja2018soft} to automatically tune the scaling parameter $\beta$ of the entropy term.
% For state value estimation, we instantiate $\mathcal{L}_V^f(y)=|\tau-\mathds{1}(y<0)| y^2$. For the entropy regularization term on Q target, we automatically tune the scaling parameter $\beta$ as in SAC~\cite{haarnoja2018soft}. 
We set the trade-off hyperparameter $\lambda$ to 0.1 in all our experiments. It might be preferable to select a larger $\lambda$ for tasks that are carried out in more reliable simulators.
% is crucially selected to balance the influence of offline real data and online simulation data on the Q-value update. 
% , and ought to be kept small if offline real data is of suboptimal quality and limited coverage or simulation has a relatively trustworthy dynamics. 
% It should be kept small when the quality and coverage of offline real data are suboptimal and limited, or when the simulation dynamics can be considered relatively reliable.

% \begin{figure*}[t]
%     \centering
%     \begin{subfigure}[b]{0.2\textwidth}
%         \centering
%         \includegraphics[width=0.95\linewidth]{./figure/Scenarios_HalfCheetah-v2.pdf}   
%         \caption{Real Environment}
%     \end{subfigure}
%     \vline
%     \hfill
%     \begin{subfigure}[b]{0.78\textwidth}
%         \includegraphics[width=0.24\linewidth]{./figure/gravity2.0.pdf} \includegraphics[width=0.24\linewidth]{./figure/torsolength4.0.pdf}  
%         \includegraphics[width=0.24\linewidth]{./figure/thighsize2.0.pdf} 
%         \includegraphics[width=0.24\linewidth]{./figure/thighsize0.5.pdf} 
%          \caption{Example environments with imperfect dynamics}
%     \end{subfigure}
%     \caption{HalfCheetah simulation environment and some of the modified dynamics.}
%     \vspace{-5mm}
%     \label{fig:simulators}
% \end{figure*}

\begin{figure*}[t]
    \begin{minipage}[t]{0.35\textwidth}
        \begin{subfigure}[t]{1.0\textwidth}
            % \vspace{-0.1cm}
            % \centering
            \includegraphics[width=0.48\linewidth]{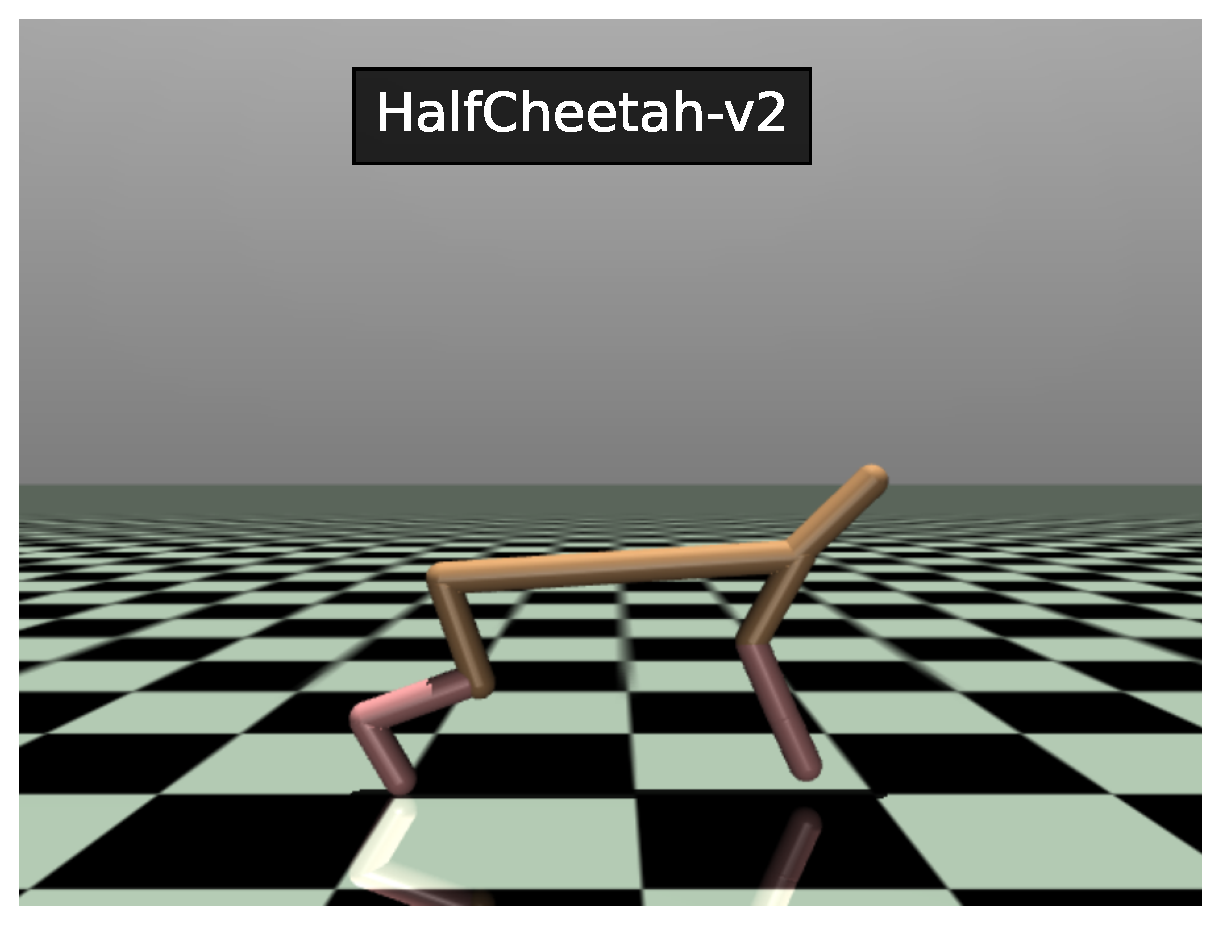} 
            \includegraphics[width=0.48\linewidth]{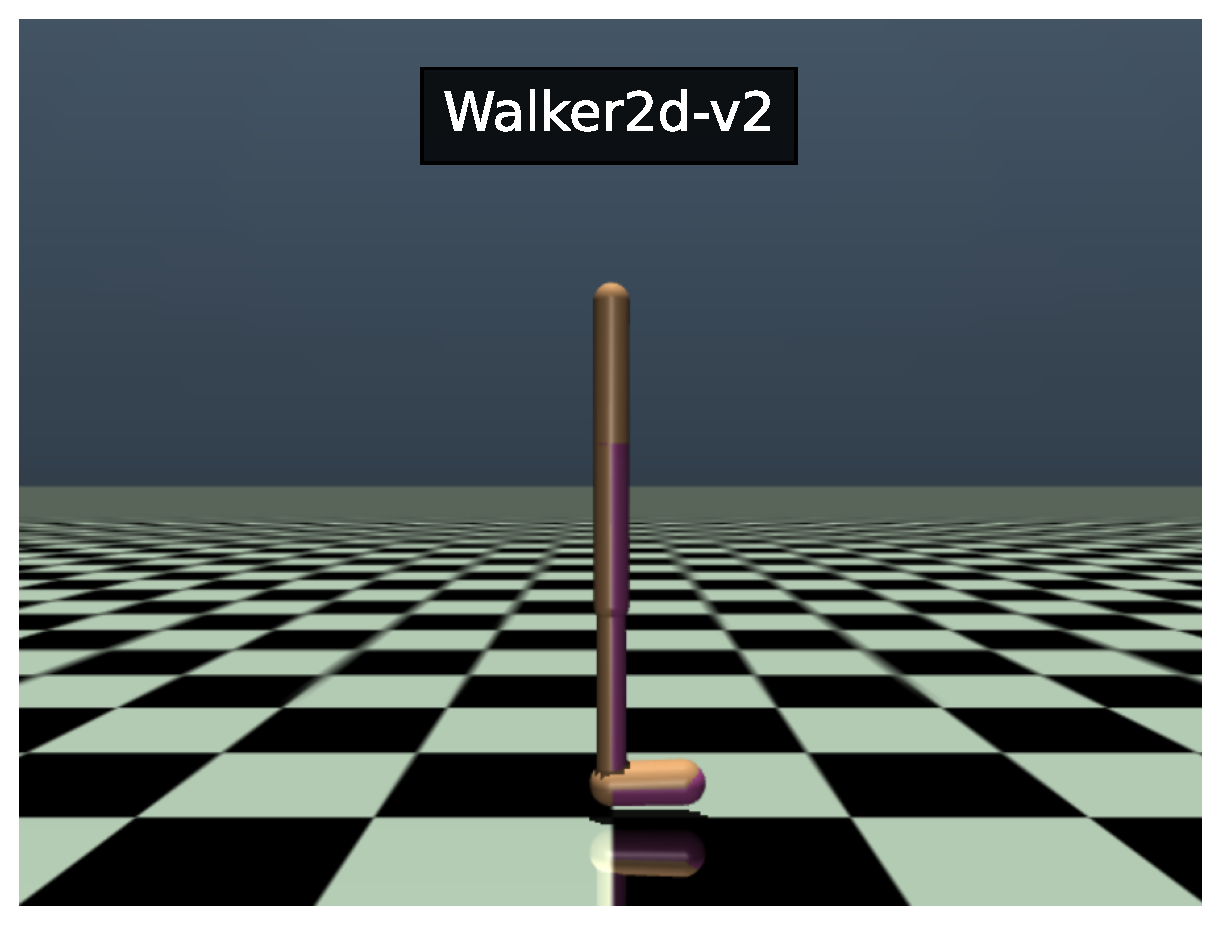} 
            \caption{Original environments}
            % \vspace{-0.3cm}
        \end{subfigure}
        \begin{subfigure}[t]{1.0\textwidth}
            % \centering
            \includegraphics[width=0.48\linewidth]{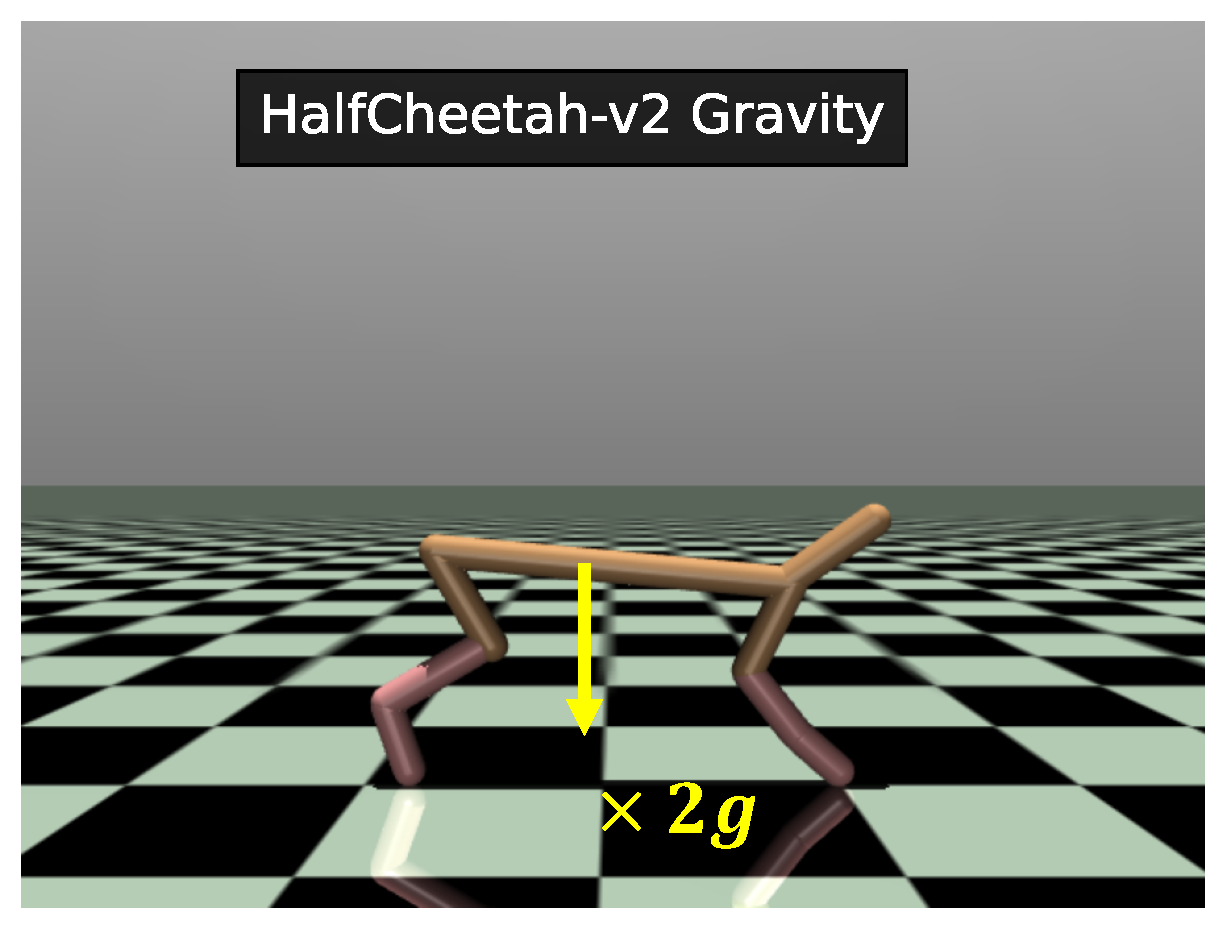}
            \includegraphics[width=0.48\linewidth]{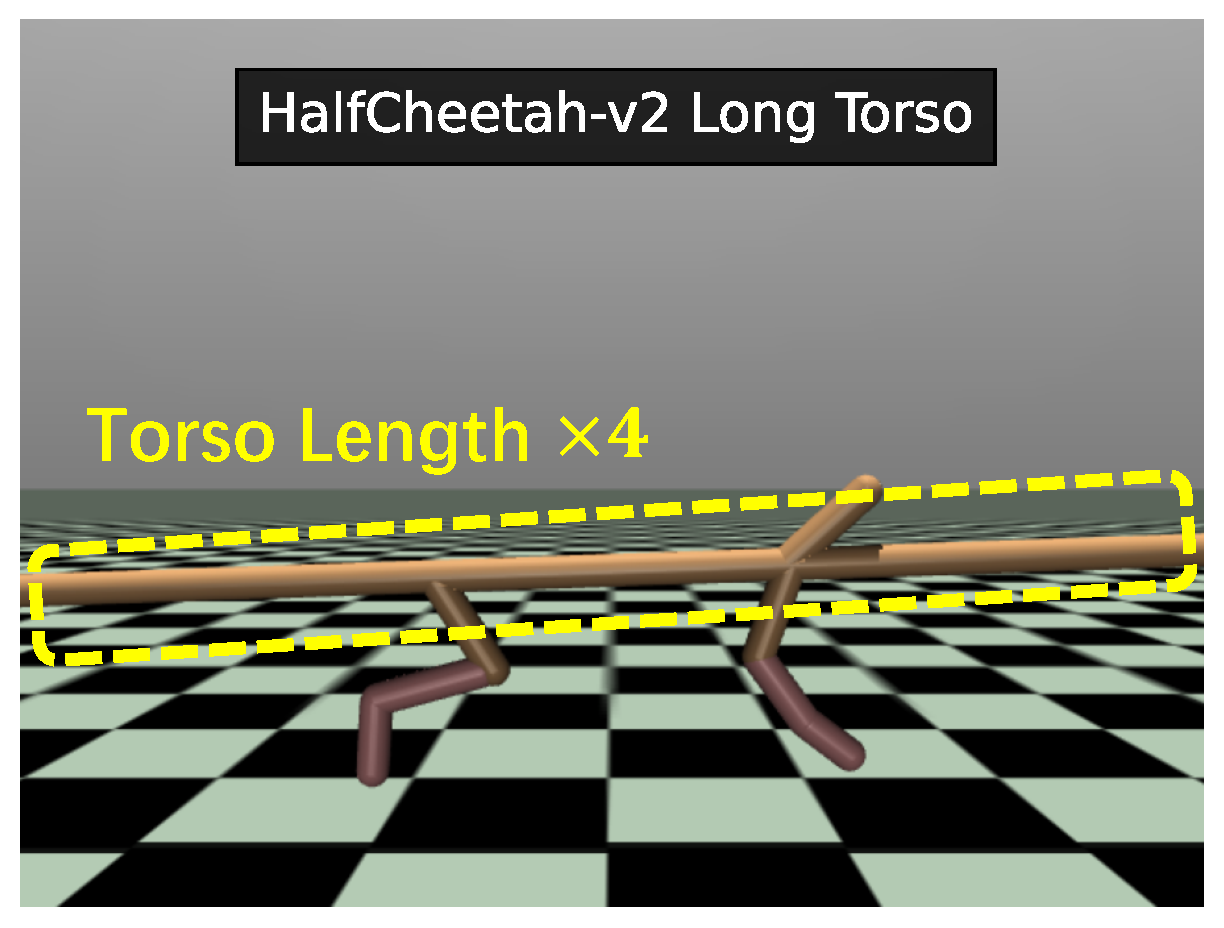}
            \vspace{0.05cm}
        \end{subfigure}
        \begin{subfigure}[t]{1.0\textwidth}
        % \centering
            \includegraphics[width=0.48\linewidth]{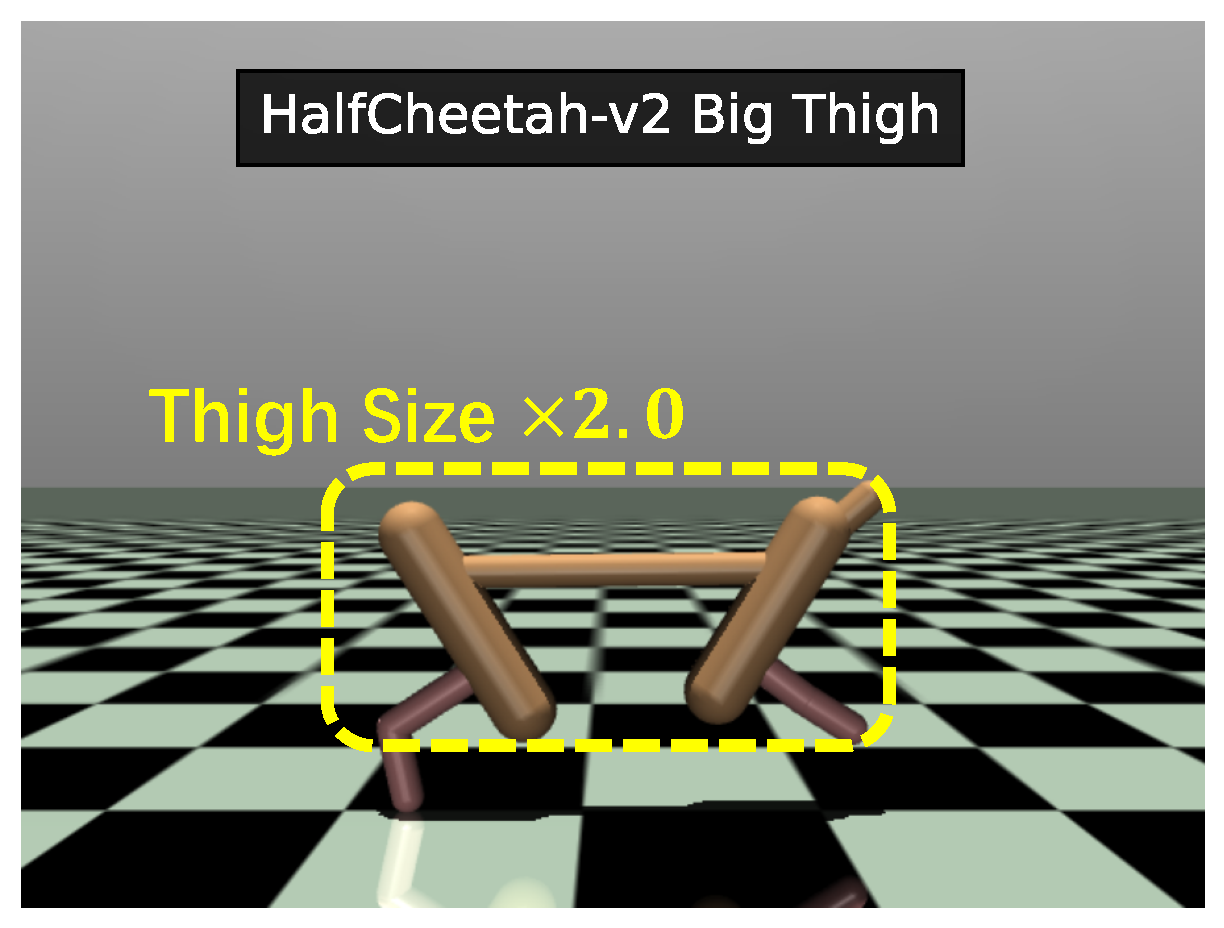} 
            \includegraphics[width=0.48\linewidth]{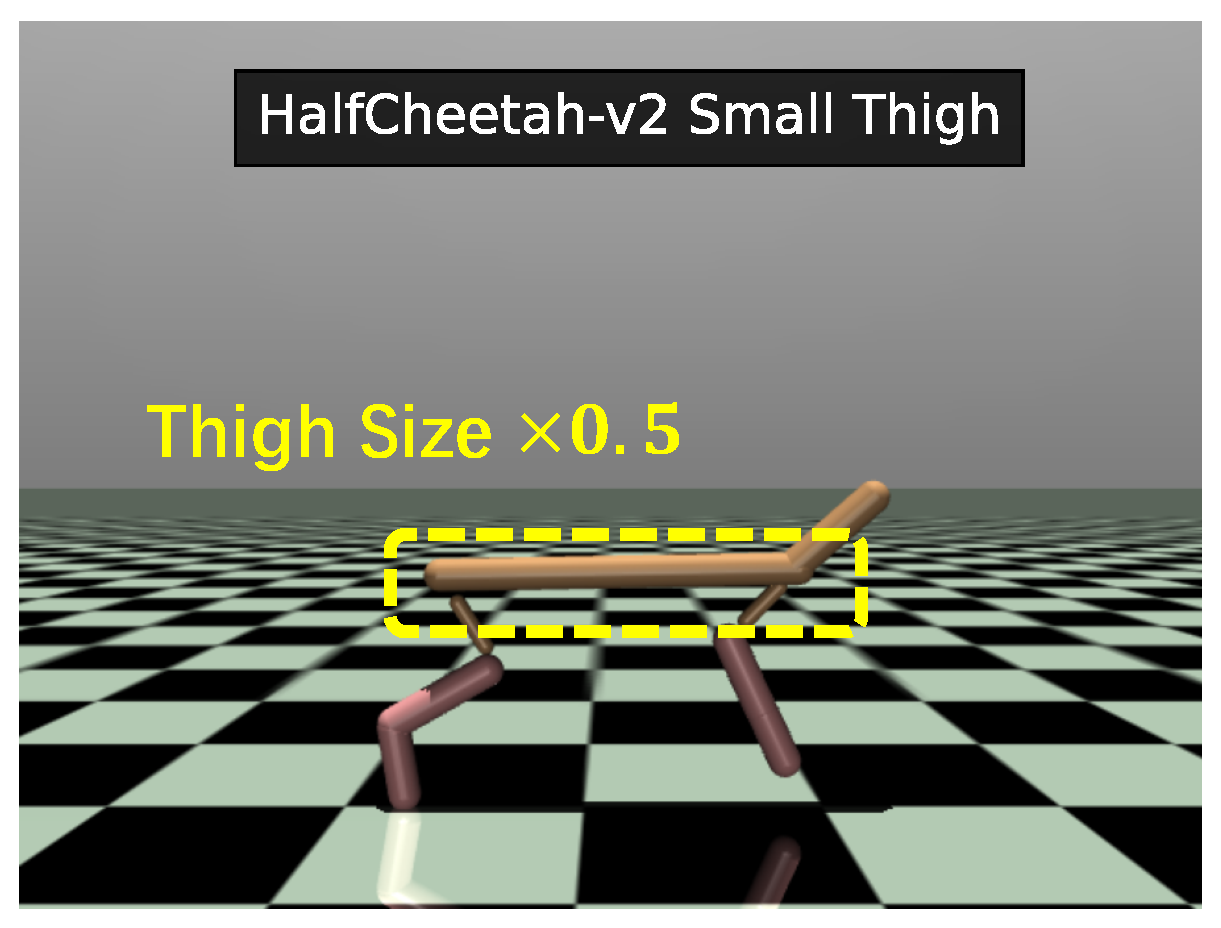} 
            \vspace{0.05cm}
        \end{subfigure}
        \begin{subfigure}[t]{1.0\textwidth}
        % \centering
            \includegraphics[width=0.48\linewidth]{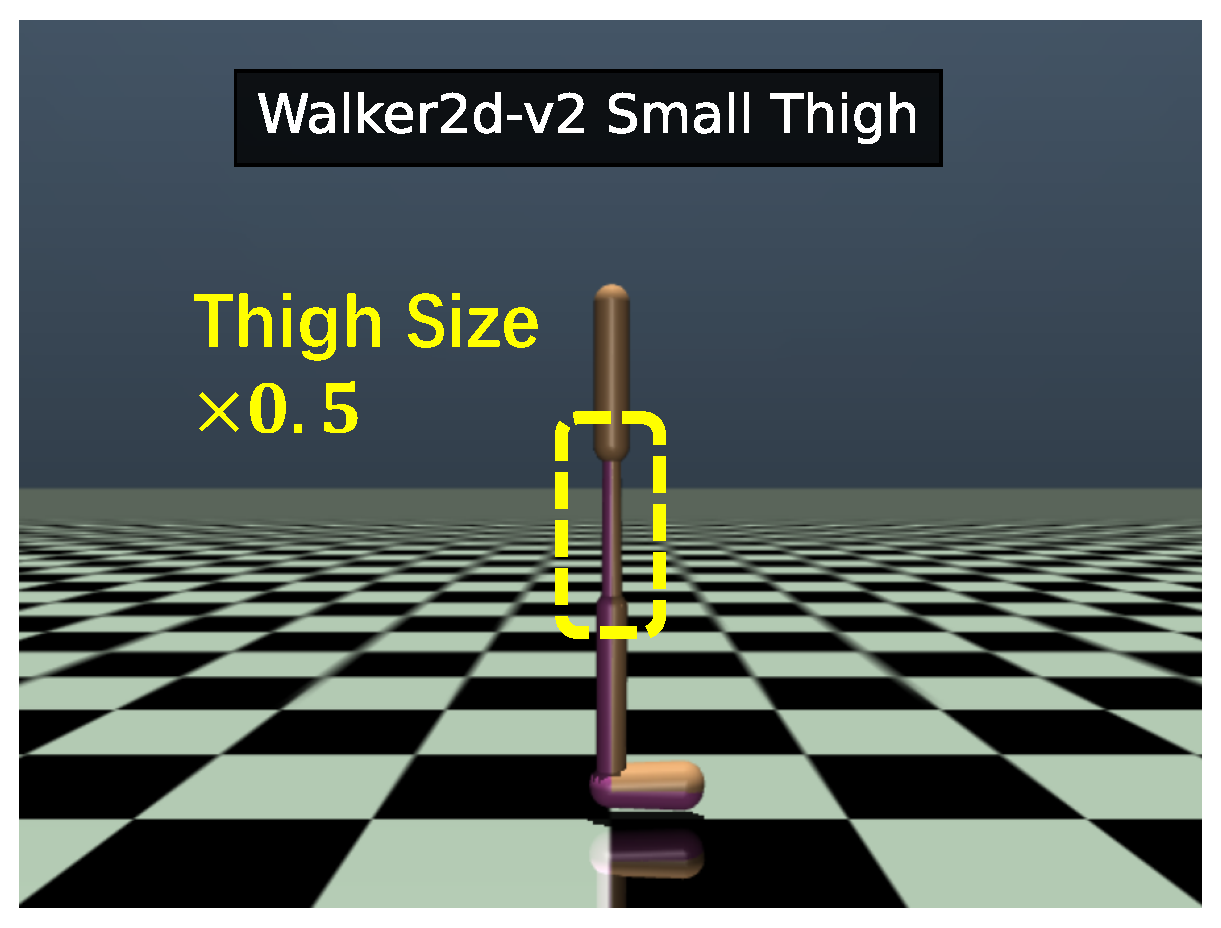} 
            \includegraphics[width=0.48\linewidth]{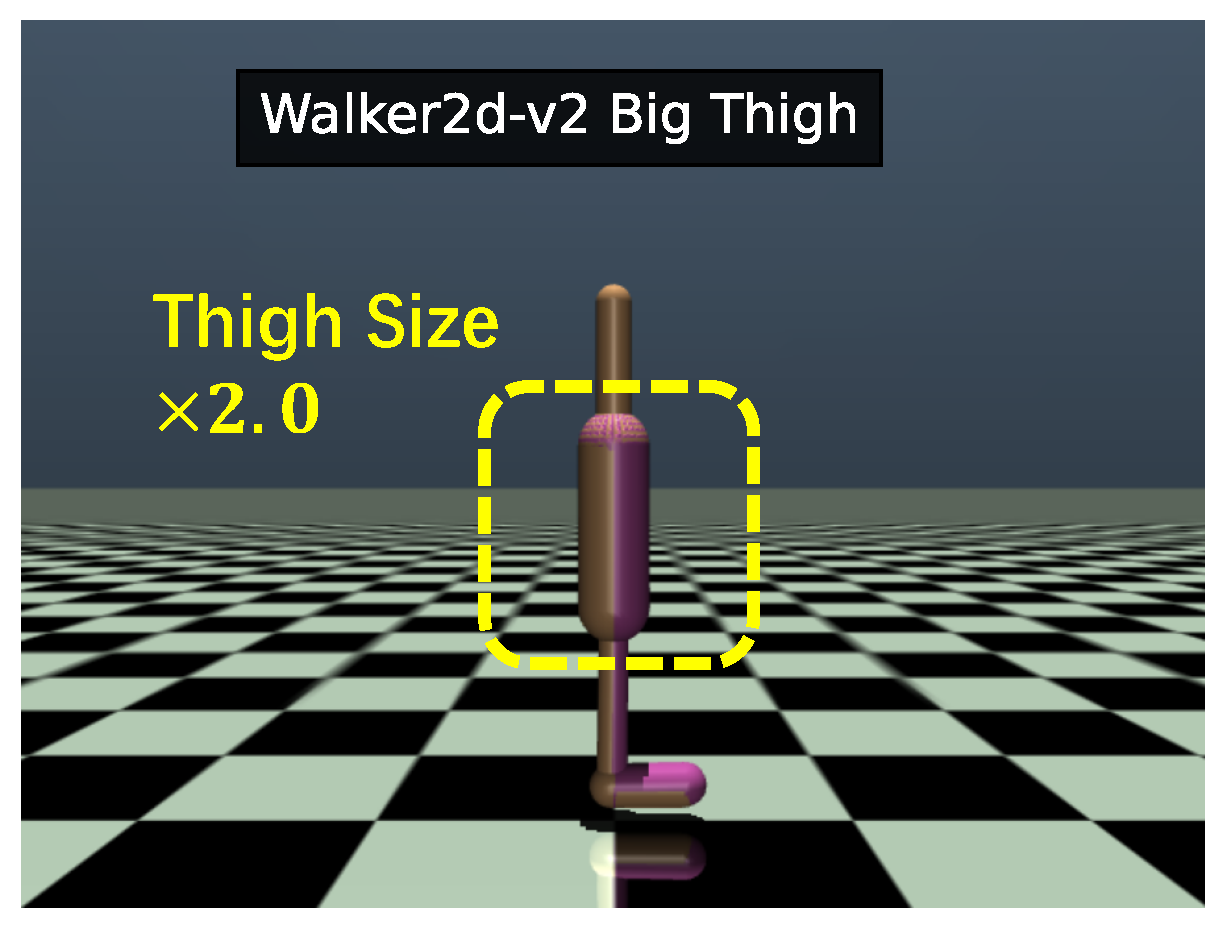} 
         \caption{Example modified environments}
        \end{subfigure}
        \caption{Original environments and some illustrations of the modified dynamics}\label{fig:simulators}
        % \vspace{-8pt}
    \end{minipage}%
    \hfill
    \begin{minipage}[t]{0.63\textwidth}
        \centering
            \vspace{-2.3cm}
            \footnotesize
            \setlength\tabcolsep{1.3pt}
            \captionsetup{type=table}
            \caption{Average returns for MuJoCo HalfCheetah and Walker2d tasks}
            \begin{tabular}{clcccccc}
            \toprule
            Data & Dynamics Gap & SAC & CQL & IQL & DARC  & H2O & H2O+
            \\\midrule

            \multirow{10}{*}{\rotatebox{90}{HalfCheetah-mr}} & Gravity & 4513$\pm$513 & 5774$\pm$214 & 5207$\pm$149 & 5105$\pm$460  & \textbf{6813$\pm$289} & \textbf{6861$\pm$268}\\
            & Friction & 2684$\pm$2646 & 5774$\pm$214 & 5207$\pm$149 & 5503$\pm$263  & 5928$\pm$896 & \textbf{6278$\pm$1336}\\
            & Joint Noise & 4137$\pm$805 & 5774$\pm$214 & 5207$\pm$149 & 5137$\pm$225 & 6747$\pm$427 & \textbf{6985$\pm$328}\\
            & Big Thigh & 4509$\pm$877 & 5774$\pm$214 & 5207$\pm$149 & 5336$\pm$389  & 6278$\pm$305 & \textbf{6675$\pm$231}\\
            & Small Thigh & 6632$\pm$1027 & 5774$\pm$214 & 5207$\pm$149 & \textbf{8331$\pm$454} & 6751$\pm$231 & 7425$\pm$148\\
            & Broken Thigh & 6517$\pm$1076 & 5774$\pm$214 & 5207$\pm$149 & \textbf{8704$\pm$1726}  & 6717$\pm$226 & 7018$\pm$147\\
            & Flexible Thigh & 5623$\pm$2862 & 5774$\pm$214 & 5207$\pm$149 & 5554$\pm$88 & 6976$\pm$234 & \textbf{7497$\pm$196}\\
            % & Ellipsoid Limb & \textbf{7853$\pm$3380} & 5774$\pm$214 & 5207$\pm$149 & 5549$\pm$43 & 5538$\pm$102 & 7040$\pm$432 & \textbf{7565$\pm$313}\\
            & Long Torso & 1047$\pm$3089 & 5774$\pm$214 & 5207$\pm$149 & 45$\pm$322  & 6225$\pm$100 & \textbf{6718$\pm$245}\\
            & Soft Feet & 5684$\pm$587 & 5774$\pm$214 & 5207$\pm$149 & \textbf{9058$\pm$374}  & 6731$\pm$319 & 7068$\pm$244\\\cmidrule{2-8}
            & \textit{Mean Return} & 4594 & 5774 & 5207 & 5863 & 6573 & \textbf{6947}\\\midrule
                \multirow{10}{*}{\rotatebox{90}{HalfCheetah-m}} & Gravity & 4513$\pm$513 & 6066$\pm$73 & 5605$\pm$25 & 5011$\pm$456  & \textbf{7085$\pm$416} & \textbf{6965$\pm$659}\\
            & Friction &  2684$\pm$2646 & 6066$\pm$73 & 5605$\pm$25 & 6113$\pm$104 & 6848$\pm$445 & \textbf{7186$\pm$859}\\
            & Joint Noise & 4137$\pm$805 & 6066$\pm$73 & 5605$\pm$25 & 5484$\pm$171 & 7212$\pm$236& \textbf{7503$\pm$237}\\
            & Big Thigh & 4509$\pm$877 & 6066$\pm$73 & 5605$\pm$25 & 6302$\pm$1832  & 6625$\pm$579 & \textbf{7094$\pm$371}\\
            & Small Thigh & 6632$\pm$1027 & 6066$\pm$73 & 5605$\pm$25 & \textbf{9127$\pm$907}  & 7020$\pm$337 & 7706$\pm$185\\
            & Broken Thigh & 6517$\pm$1076 & 6066$\pm$73 & 5605$\pm$25 & \textbf{7509$\pm$707}  & 6800$\pm$378 & 7321$\pm$213\\
            & Flexible Thigh & 5623$\pm$2862 & 6066$\pm$73 & 5605$\pm$25 & 7266$\pm$1771  & 7005$\pm$757 & \textbf{7805$\pm$139}\\
            & Long Torso & 1047$\pm$3089 & 6066$\pm$73& 5605$\pm$25 & 724$\pm$921  & \textbf{6327$\pm$602} & 5484$\pm$1382\\
            & Soft Feet & 5684$\pm$587 & 6066$\pm$73 & 5605$\pm$25 & 6952$\pm$3330  & 7138$\pm$326 & \textbf{7622$\pm$53}\\\cmidrule{2-8}
            & \textit{Mean Return} & 4594 & 6066 & 5605 & 6054 & 6896 & \textbf{7187}\\\midrule\midrule
                    \multirow{8}{*}{\rotatebox{90}{Walker2d-mr}} & Gravity   & 1698$\pm$1611 & 3261$\pm$802 & 3390$\pm$326 & 2969$\pm$1043 & 3366$\pm$740 & \textbf{3518$\pm$605}                \\ %\cline{2-7} 
                      & Friction   & 2779$\pm$870 & 3261$\pm$802 & 3390$\pm$326 & 3644$\pm$213 & \textbf{3916$\pm$549} &  \textbf{3866$\pm$840} \\ %\cline{2-7} 
                      & Joint Noise& 173$\pm$727 & 3261$\pm$802 & 3390$\pm$326 & -3$\pm$0 & 3045$\pm$911 &  \textbf{3446$\pm$862}  \\ 
                      & Big Thigh& 1151$\pm$716 & 3261$\pm$802 & \textbf{3390$\pm$326} & 57$\pm$126 & 1789$\pm$1781 &  2977$\pm$771  \\
                      & Small Thigh& 894$\pm$519 & 3261$\pm$802 & 3390$\pm$326 & 1294$\pm$905 & 2455$\pm$1301 &  \textbf{3920$\pm$417}  \\
                       & Broken Thigh& \textbf{3845$\pm$607} & 3261$\pm$802 & 3390$\pm$326 & 893$\pm$180 & 2702$\pm$1054 &  \textbf{3911$\pm$405}  \\
                        & Flexible Thigh& 2518$\pm$1627 & 3261$\pm$802 & 3390$\pm$326 & 2511$\pm$1048 & 1891$\pm$1001 &  \textbf{3535$\pm$493}  \\\cmidrule{2-8}
                        & \textit{Mean Return}& 1865 & 3261 & 3390 & 1624 & 2738 &  \textbf{3596}  \\
            \bottomrule
        \end{tabular}\label{tab:halfcheetah}
    % \end{table}
    \end{minipage}
\vspace{-8pt}
\end{figure*}

\subsubsection{Simulation experiments}
% 	\subsubsection{Simulation Experiment Setup}\label{sim_setup}
%	For the simulation-based experiments, 
% We conduct simulation-based experiments in the MuJoCo physics simulator~\cite{todorov2012mujoco}. In particular, we construct ten new simulation task environments (serve as the simulated environments) with intentionally introduced dynamics gaps upon the original MuJoCo-HalfCheetah task environments (serve as the real environments) by modifying the dynamics parameters, list as follows:
We treat the original MuJoCo task environment as the ``real-world'' scenario, and create ten imperfect simulation environments by deliberately introducing various types of dynamics gaps (illustrated in Figure~\ref{fig:simulators}).
% We create ten imperfect simulation environments based on the MuJoCo physics simulator with deliberately introduced dynamics gaps. We treat the original MuJoCO
These dynamics gaps are introduced by adjusting either the dynamics parameters of the robot or the environmental physical properties. 
For example, in the HalfCheetah and Walker2d task environment, 
% These imperfect simulators are derived from the original MuJoCo-HalfCheetah task environment, which acts as our real-world scenarios. As illustrated in Figure~\ref{fig:simulators}, 
% the dynamics gaps are introduced by adjusting either the dynamics parameters of the robot or the environmental physical properties, 
the modifications include modifying the gravitational gravity ($\times$2, \textbf{Gravity}), friction coefficient ($\times$0.3, \textbf{Friction}), thigh size ($\times$0.5 and $\times$2, \textbf{Small/Big Thigh}),
% \begin{wrapfigure}{r}{.4\textwidth}
% \end{wrapfigure}
the motion range of the joint connections of thighs ($\times$0.5 and $\times$2, \textbf{Broken/Flexible Thigh}), stretching the torso length ($\times$4, \textbf{Long Torso}, only for HalfCheetah), lowering the foot stiffness ($\times$0,\textbf{ Soft Feet}, only for HalfCheetah) and adding joint noise ($N(0,\mathbf{I})$, \textbf{Joint Noise}).
In terms of the ``real-world'' offline dataset (original MuJoCo environment), we utilize the corresponding task datasets from the widely-used offline RL benchmark D4RL~\cite{fu2020d4rl}. 
Specifically, the Medium (-m) and Medium Replay (-mr) datasets are considered as they are closer to real-world settings, 
where we are more likely to obtain medium-level or highly mixed offline datasets from real systems.
% since they reflect representative real-world settings where it is considerably impractical to acquire datasets of high quality and broad coverage. 
The online training of algorithms is performed in the created imperfect simulation environment, and we evaluate the learned policy in the original MuJoCo environment.
% Online training with offline data takes place in our created simulation environments, while the evaluation of the learned policy performance is conducted in the original untouched HalfCheetah environment. 
% We visualize some of our constructed simulators and the corresponding real environment in Figure~\ref{fig:simulators}.

% \vspace{4pt}
% \begin{figure}[t]
%     \centering
%     \begin{subfigure}[b]{0.49\textwidth}
%         \centering
%         \includegraphics[width=\textwidth]{figure/standing_dataset.pdf}
%         \caption{Standing still dataset}
%         \label{fig:standing_dataset}
%     \end{subfigure}
%     \begin{subfigure}[b]{0.49\textwidth}
%         \centering
%         \includegraphics[width=\textwidth]{figure/moving_dataset.pdf}
%         \caption{Moving forward dataset}
%         \label{fig:moving_dataset}
%     \end{subfigure}
%     \caption{State, action and reward distributions of the datasets}
% \end{figure}

\vspace{4pt}
\subsubsection{Real-robot transfer experiments}
We also perform real robot transfer experiments on a wheel-legged robot with a main body and a pair of legs with wheels attached to the end. We also construct the simulation environment based on Isaac Gym~\cite{makoviychuk2021isaac}. Both the real robot and its simulation are shown in Figure~\ref{fig:robot}.
Our wheel-legged robot bears a substantially large weight (about 12 kg) and possesses an intricate mechanical structure. Moreover, our testing environment features a furry carpet on the ground, which introduces the possibility of sagging and unloading as the robot traverses the surface. These distinctive factors collectively contribute to a very challenging real-world transfer procedure.
% that balance itself on its two wheels at the end of its legs. 
% We also evaluate the performance of H2O in a real wheel-legged robot control scenario. 
% As shown in Figure, the robot consists of a main body and a pair of legs with wheels attached to the end. The main body is composed of a chassis, a on-board micro-computer, a battery and two sensors. There are eight actuators in total, including six leg joint motors and two wheel motors, as illustrated in Figure. It is also notable that the total weight of the robot reaches at 12 kilograms.

According to its sensors and actuators, the state space of the robot control tasks is designed as a quadratic-tuple $(\theta, \dot\theta, x, v)$ where $\theta$ denotes the forward pitch angle of the body, $x$ is the displacement of the robot, $\dot\theta$ and $v$ are the angular and linear velocity respectively. The execution action is the torque $\tau$ of the motors at the two wheels. We construct two tasks for real-world validation: (1) \textbf{standing still}: the robot needs to keep balanced at the initial location and maintain stability as much as possible. The reward $r$ is calculated as: $ r=30.0- x^{2}- {v}^{2}-\theta^{2} - {\dot \theta}^{2}-\tau^2$. Ideally, when the robot standing at the original space without any swinging, the penalty item $x^{2}+v^{2}+\theta^{2}+{\dot \theta}^{2} $ will be minimized. To further avoid shaking and to protect the wheel motor, we limit the wheel torque by adding a penalty $\tau^2$ into the reward.
%	state to $[0, 0, 0, 0]^T$; 
(2) \textbf{moving forward}: the robot needs to move at a fixed forward speed $v_{tgt}=0.2 m/s$ and maintain balance as long as possible.
The state space of this task $\mathbf{s} =(v, \theta, \dot \theta)$, where the state $x$ is no longer included as we hope that the robot can move forward in any displacement. The reward is calculated as:
$r=15.0- {({v}-0.2)}^2-\tau^2$, in which we use the penalty item ${(v-0.2)}^2$ to regulate the moving speed of the robot while adding $-\tau^2$ to avoid shaking.
We collect offline datasets for the standing still and moving forward tasks respectively by recording 16,588 real-world human-controlled transitions (about 90s of real-time control). 
\vspace{-4pt}
\subsection{Comparative Results}
\subsubsection{Simulation experiments} The results presented in Table~\ref{tab:halfcheetah} highlight the superiority of H2O+ compared to all the baseline methods in terms of the mean return across all tasks in the HalfCheetah and Walker2d environments. Note that for offline RL baselines (CQL and IQL), we train the policies using the "real" offline datasets and evaluate them in the "real" environments, so their scores remain the same across modified environments with dynamics gaps.

% Notably, H2O+ demonstrates comparable performance to offline RL\footnote{The results of offline RL baselines keep consistent on the same offline dataset adopted, since they aren't involved into online simulation explortaion.} when online zero-shot transfer struggles to manage, and vice versa, showcasing its robustness to biased simulation samples or suboptimal real-world datasets. 
It is found that online cross-domain baseline DARC performs strongly when the dynamics gap is small (i.e. when online SAC achieves better performance than offline RL baselines). However, DARC fails miserably when the dynamics gap is large (e.g. HalfCheetah long torso and Walker2d joint noise tasks), which underscores its limitations of the dynamics-gap-related reward penalization scheme.
On the other hand, offline RL baselines are not impacted by the sim-to-real issue, but their performances are heavily impacted by the dataset quality. Our proposed H2O+ not only outperforms H2O in most tasks, but also consistently achieves comparable or better performance than online, offline, and cross-domain RL methods in both small and large dynamics gap settings. These results demonstrate the effectiveness of H2O+ in leveraging both offline data and imperfect simulation for improved and transferable policy learning.

\vspace{4pt}
\subsubsection{Real-robot experiments}
\begin{figure*}[t]
    \centering
        \begin{subfigure}[b]{0.3\textwidth}
            \centering
            \includegraphics[height=2.6cm,keepaspectratio]{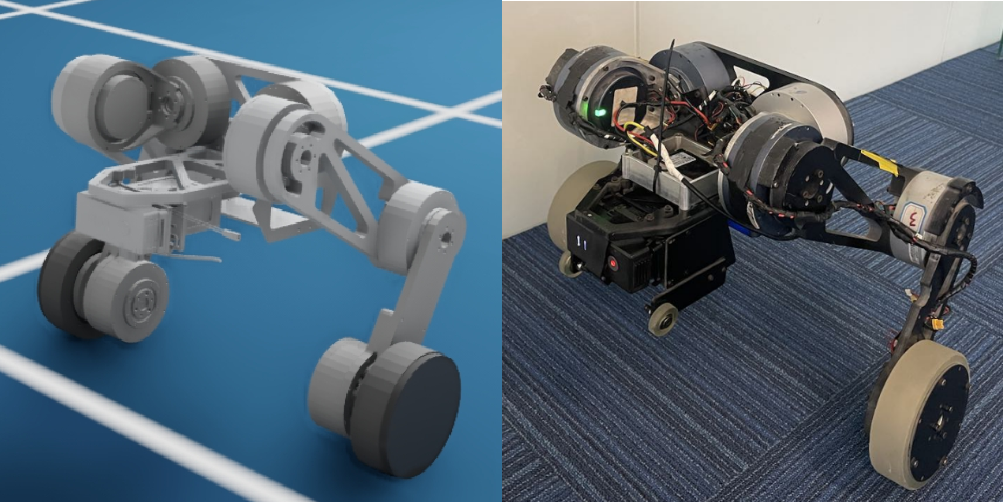}
            \caption{Simulation/Real robot}
            \label{fig:robot}
        \end{subfigure}
        \hfill
        \begin{subfigure}[b]{0.34\textwidth}
            \centering
        \includegraphics[height=3.0cm,keepaspectratio]{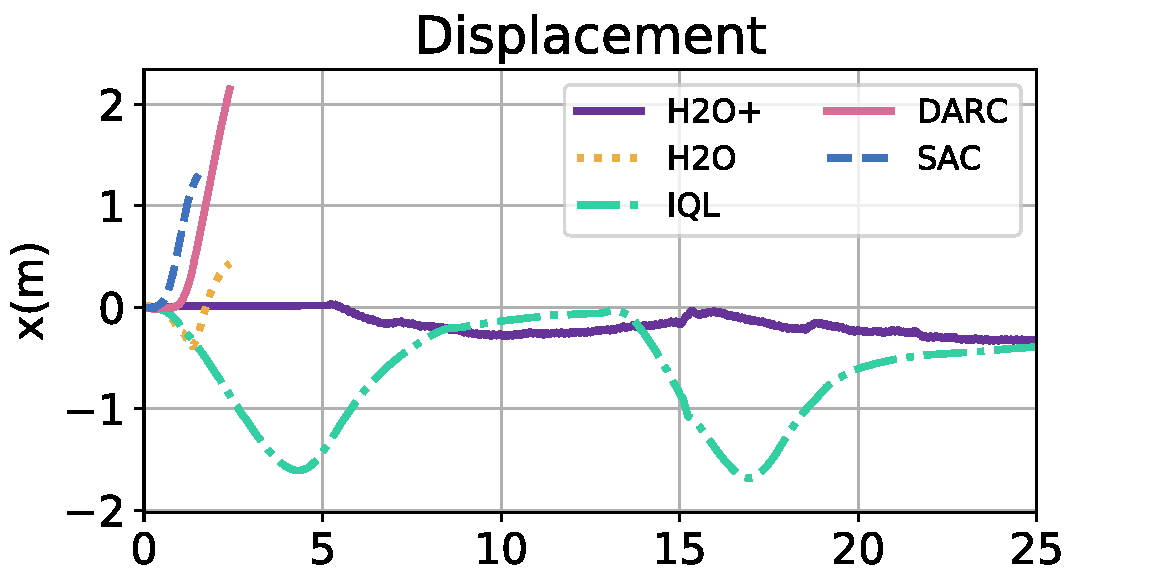}
            \caption{Displacement in standing still}
            \label{fig:real_result_main_a}
        \end{subfigure}
        \hfill  
        \begin{subfigure}[b]{0.34\textwidth}
            \centering
            \includegraphics[height=3.0cm,keepaspectratio]{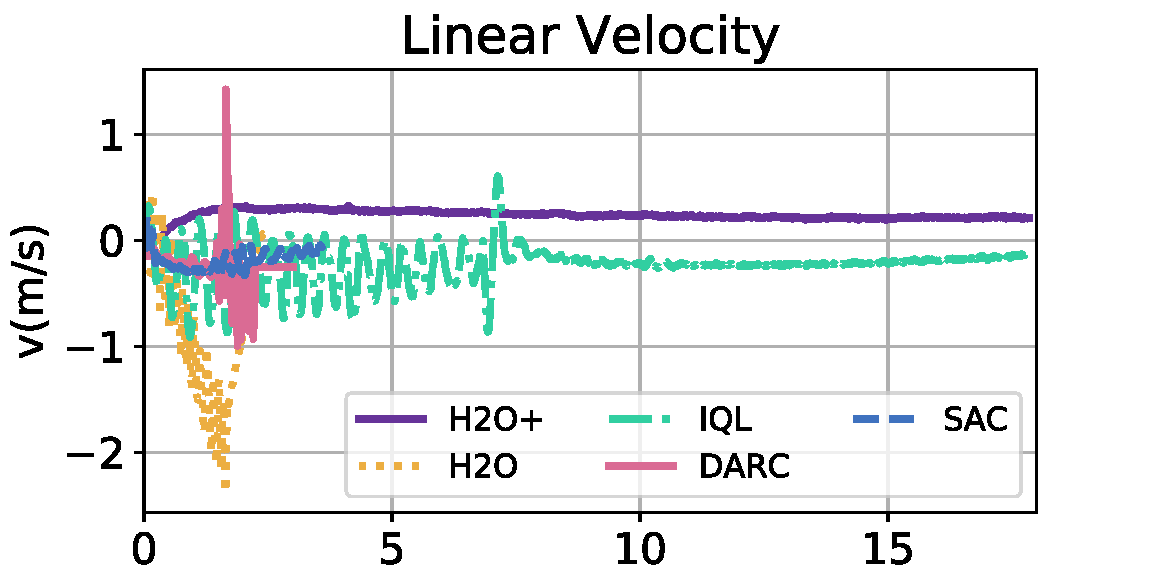}
            \caption{Linear velocity in moving forward}
            \label{fig:real_result_main_c}
        \end{subfigure}
        \vspace{-4pt}
        \caption{The real-robot experiment results of the “standing still” (b) and “moving forward” (c) tasks}
% (a) Displacement. (b) Simulated data coverage and normalized
% rewards (illustrated in color). Results of the “moving forward” task: (c) Linear
% velocity, (d) Simulated data coverage and normalized rewards.}
\label{fig:real_result_main}
    % \caption*{}
    \vspace{-12pt}
\end{figure*}

\begin{figure}[t]
    \centering
    \begin{subfigure}[b]{0.46\textwidth}
        \centering
        \includegraphics[width=\textwidth, keepaspectratio]{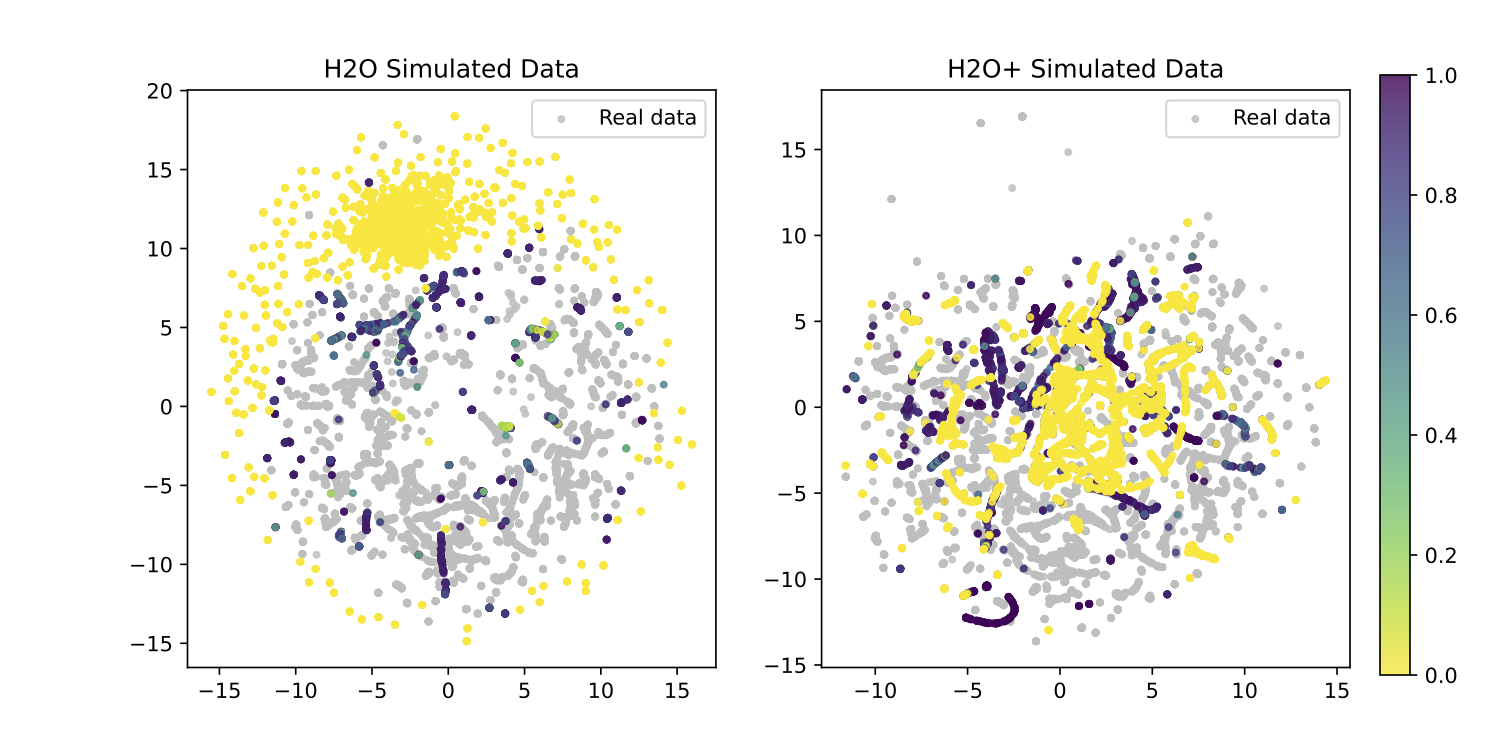}
        \caption*{}
        \label{fig:real_result_main_b}
    \end{subfigure}
    
    \vspace{-8mm}
    \begin{subfigure}[b]{0.46\textwidth}
        \centering
        \includegraphics[width=\textwidth, keepaspectratio]{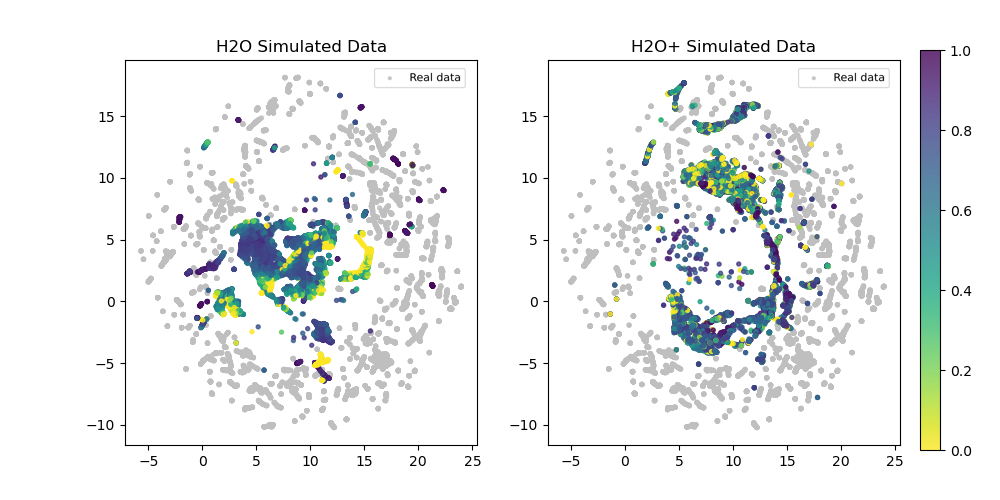}
        \caption*{}
        \label{fig:real_result_main_d}
    \end{subfigure}
    \vspace{-22pt}
    \caption{Comparison of H2O / H2O+ simulation data quality in real-world tasks. (Top: standing still; Down: moving forward)}
    \label{fig:horizontal_subfigures}
    \vspace{-4pt}
\end{figure}

    In the real-robot experiments, H2O+ demonstrated a strong transfer ability compared to other benchmarks. As shown in Figure~\ref{fig:real_result_main_a} and~\ref{fig:real_result_main_c}, the control performance of this method far exceeds that of others in both tasks. In the \textbf{standing still} task, only H2O+ and IQL policies successfully maintain the balance of the robot for over 30 seconds (s), while the robot deployed with SAC, H2O, and DARC policies cannot even keep the balance for over 3s before it hit the ground. Moreover, as illustrated in Figure~\ref{fig:real_result_main_a}, H2O+ regulated the displacement of the robot within 0.2m, whereas IQL only barely maintains balance, yet with a large range swinging even reaches 1.6m. The advantage of H2O+ is more significant in the \textbf{moving forward} task. As illustrated in Figure~\ref{fig:real_result_main_c}, only the H2O+ policy achieves the goal of moving forward and even follows the target velocity precisely. During the moving process, the speed of robot changes smoothly and the pitch angle remains steady. In comparison, IQL policy is capable to keep balance, but the robot moves backward and also spends more time and effort on keeping balance, resulting in a shaking period of over 7s. In addition, H2O, SAC and DARC fail to maintain balance and fall down in 4s. Additionally, as in Figure~\ref{fig:horizontal_subfigures}, we observe that H2O explores a more focused high-value area, whereas H2O+ spans a broader high-value region, thus indicating superior diversity characteristics in simulated data, which would benefit the overall performance. We also test H2O+ in a more biased simulator with an intentionally introduced mass reduction, while H2O+ still outperforms other baselines. Please see the \href{https://sites.google.com/view/h2oplusauthors/}{H2O+ webpage} for more details.

%     We further investigate on the quality of online interactions explored by H2O+ compared to H2O, from the data coverage and the value. In Figure~\ref{fig:coverage-standingstill} and Figure~\ref{fig:coverage-movingstraight}, we visualize the coverage and the normalized value of displacement, velocity, angle, angular velocity, and action in the real-world robot task, ``standing still'' and ``moving forward'' respectively. Specifically, we also visualize the real data~(collected offline in the real environment) with gray points in these two figures, and the colorful points reveal the data collected in the simulator. 

% In the ``standing still'' task, we observe that H2O explores a more focused high-value area, whereas H2O+ spans a broader high-value area, thus demonstrating superior diversity characteristics in simulated data, which would benefit the overall performance.

% For the task of ``moving forward'', a clear distinction can be observed between the quality of the simulated data gathered by H2O and H2O+. Notably, the H2O+ approach excels in terms of data coverage, meaning it successfully spans a broader region of the state-action space. Additionally, the data collected by H2O+ exhibits better dispersion compared to H2O's data, indicating a higher degree of diversity. This superior diversity, in turn, contributes to a more comprehensive exploration of the state-action space and enhances the robustness of the learned policy.

\vspace{-3mm}
\subsection{Ablation Studies}
\vspace{-1mm}

\subsubsection{Ablation on the hyperparameter $\lambda$ and dynamics ratio}
Table~\ref{tab:ablation} presents the results of different choices of the trade-off parameter $\lambda$ and the dynamics ratio on Bellman error in dynamics-aware mixed 
value update, as used in Eq.~\ref{eq:final_Q}. Specifically, we investigate the cases where $\lambda=0$ corresponds to Q-value update on both simulation and real data using Eq.~\ref{eq:evaluation_Q}, and $\lambda=1$ formulates Q-value update with Eq.~\ref{eq:insample_Q}. 
Among the different parameter choices, the original implementation of H2O+ with a mixed Q-value update at $\lambda=0.1$ achieves the highest performance compared to other selections. 
It reveals that there is no necessity to heavily regulate the Q target by incorporating too much information from the value function learned as an anchor from offline data. 
Moreover, it is evident that H2O+ exhibits a remarkable level of hyper-parameter insensitivity, as indicated by its minor performance discrepancies across the $\lambda$ range of 0.0 to 0.5.
Essentially, our analysis also reveals that the absence of dynamics ratio reweighting in the Bellman error results in significant performance degradation.

% \begin{minipage}{0.4\textwidth}

% \end{minipage}
% \begin{minipage}{0.4\textwidth}

% \end{minipage}

\begin{figure}[t]
    \vspace{-8pt}
    \centering
    \includegraphics[width=0.42\textwidth]{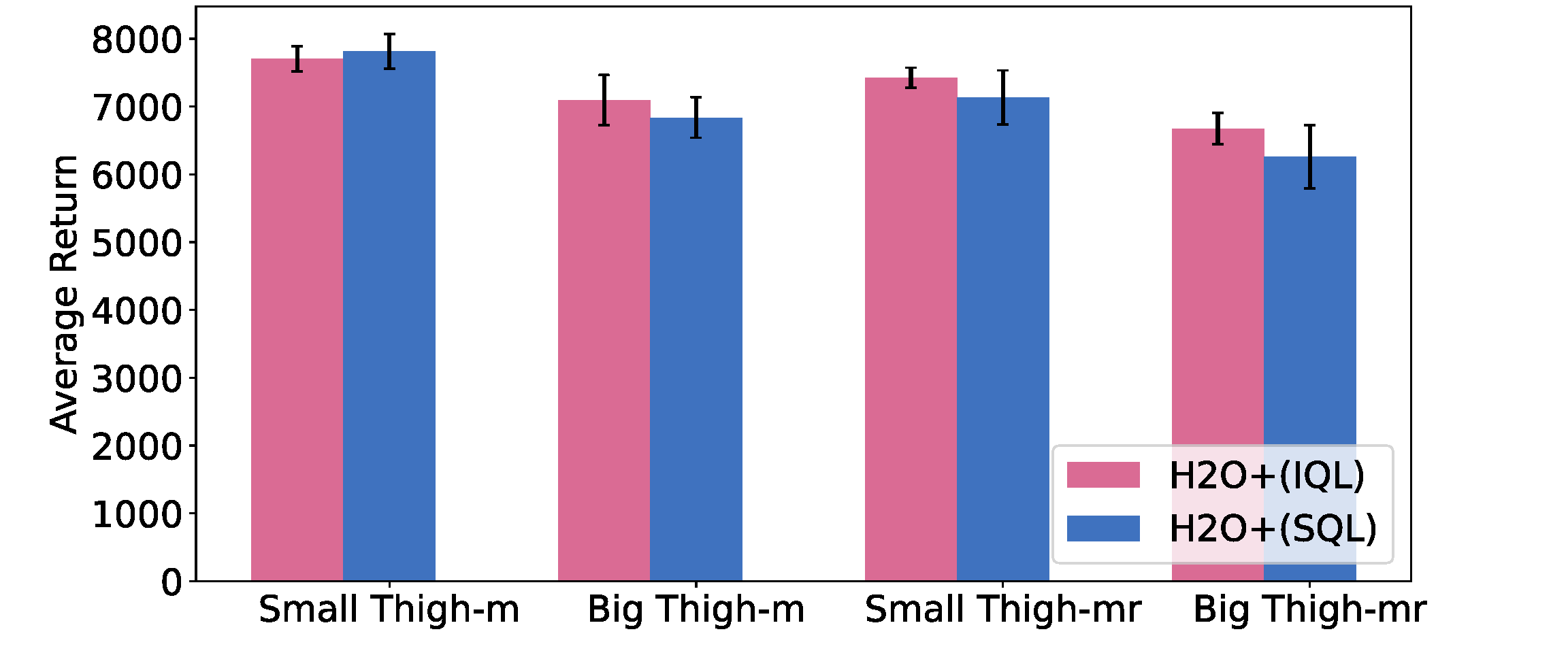}
    \vspace{-3pt}
    \caption{Different choices of offline RL backbone for state-value function learning}
    \label{fig:intro_optimal_gap}
    \vspace{-12pt}
\end{figure}

\begin{table}[t]
\centering
    % \vspace{-8pt}
    \caption{Ablations on choices in dynamics-aware mixed value update designings ($\lambda$ and dynamics ratio)}
    % \scriptsize
    \setlength\tabcolsep{4pt}
    % \adjustbox{center}{
    \begin{tabular}{c|ccc}
        \toprule
          % & \multicolumn{4}{c}{H2O+} \\\midrule
         Trade-off $\lambda$ & 0.0 & 0.1 & 0.2\\
         Dynamics ratio & \cmark & \cmark & \cmark\\
        Average return &  6738$\pm$444  &  \textbf{6861$\pm$268}  & 6677$\pm$252  \\\midrule
        Trade-off $\lambda$ & 0.5 & 1.0 & 0.1 \\
         Dynamics ratio & \cmark & \cmark & \xmark \\
        Average return & 6563$\pm$752 & 6242$\pm$68 &  5579$\pm$530  \\
        % Updated one with 5 seeds
      %  Average Return & \textbf{6642$\pm$324} & 6568$\pm$194 & 4866$\pm$303 & 5381$\pm$406 & 6396$\pm$175 & 5322$\pm$345 & 5947$\pm$59 & 5001$\pm$548 \\
    \bottomrule
    \end{tabular}
    % }
    \label{tab:ablation}
\vspace{-10pt}
\end{table} 

\begin{table}[t]
\centering
    \caption{Ablations on different levels of dynamics gap}
    % \scriptsize
    \setlength\tabcolsep{4pt}
    % \adjustbox{center}{
    \begin{tabular}{c|cccc}
        \toprule
          % & \multicolumn{4}{c}{H2O+} \\\midrule
         Gravity &  @1.25 & @1.5 & @2.0 & @3.0 \\\midrule
         H2O & 6846$\pm$572 & 6483$\pm$529 & 6813$\pm$289 & \textbf{6171$\pm$1209} \\
        H2O+ & \textbf{7165$\pm$134}  & \textbf{6948$\pm$258} & \textbf{6861$\pm$268} & \textbf{6135$\pm$811}  \\
        % Updated one with 5 seeds
      %  Average Return & \textbf{6642$\pm$324} & 6568$\pm$194 & 4866$\pm$303 & 5381$\pm$406 & 6396$\pm$175 & 5322$\pm$345 & 5947$\pm$59 & 5001$\pm$548 \\
\bottomrule
    \end{tabular}
    % }
    \label{tab:diff_gravity_acc}
    \vspace{-10pt}
\end{table} 

\vspace{4pt}
\subsubsection{Ablation on offline RL backbones}
Furthermore, we plug in other offline RL backbones like SQL~\cite{xu2023offline} 
into H2O+ paradigm for state value learning, 
by replacing the value loss in Eq.(\ref{eq:offline_J}) with $\mathcal{L}_V^f(y)=\mathds{1}(1+y/2\alpha>0)(1+y/2\alpha)^2-y/\alpha$. 
We demonstrate H2O+ offers flexibility to bridge other offline RL algorithms in Small and Big Thigh tasks on HalfCheetah Medium and Medium Replay datasets, producing comparable performance as shown in Figure~\ref{fig:intro_optimal_gap}.

% We also ablate on $\lambda$ choices in the mixed bellman operator on HalfCheetah-mr gravity task. It reveals that there is no necessity to heavily regulate the Q target by incorporating too much information from the value function learned as an anchor from offline data. Moreover, it is evident that H2O+ exhibits a remarkable level of hyper-parameter insensitivity, as indicated by its minor performance discrepancies across the $\lambda$ range of 0.0 to 0.5.

\vspace{4pt}
\subsubsection{Investigations on different levels of dynamics gaps}
We further compare H2O and H2O+ under different levels of dynamics gaps (HalfCheetah-mr with 1.25 to 3 times gravity). The results are presented in Table~\ref{tab:diff_gravity_acc}. It is observed that H2O+ beats H2O in all different dynamics discrepancy levels, despite using a simpler approach to handle dynamics gaps. 
% Moreover, H2O+ achieves smaller variance in every task, indicating that it is more stable than H2O. Furthermore, H2O+ performs much better in low dynamics gap tasks and even doesn't lose in high dynamics ones, echoing the designing philosophy of unleashing potentials of online simulation with less conservatism.
In addition, H2O+ consistently demonstrates a lower variance across all tasks, underscoring its heightened stability compared to H2O. Furthermore, H2O+ performs much better in tasks with low dynamics gaps and still maintains competitive performance in high dynamics scenarios. This alignment with the underlying design philosophy of leveraging the full potential of online learning with less conservatism further accentuates the superiority of H2O+.

\section{Conclusion}\label{conclusion and Future Directions}
In this paper, we propose an improved hybrid offline-and-online RL framework (H2O+) to enable full utilization of real-world offline datasets and imperfect simulators for cross-domain policy learning. Our method addresses several key weaknesses in the previous method H2O, and offers flexibility to bridge various choices of strong offline RL backbones without introducing excessive conservatism. Through extensive simulation and real-world experiments, we show that our method outperforms the SOTA cross-domain RL methods in a wide range of dynamics gap settings. This makes H2O+ an ideal candidate for many real-world tasks without high-fidelity simulators and sufficient offline data.

\bibliographystyle{IEEEtran}
\bibliography{root}  % .bib

\newpage

\end{document}